\documentclass[journal]{IEEEtran}

\usepackage{times}

\usepackage{microtype}

\usepackage{caption}
\usepackage{amsmath}
\usepackage{amsfonts}
\usepackage{mathabx}
\usepackage{xcolor}
\usepackage[capitalize]{cleveref}
\usepackage{enumitem}
\usepackage{multirow}
\usepackage{graphicx}
\usepackage[mathscr]{euscript}
\usepackage{makecell}
\usepackage{soul}

\DeclareMathOperator*{\argmax}{arg\,max}

\let\vec\mathbf

\ifCLASSINFOpdf
\else
\fi
\usepackage{url}

\begin{document}
%
\title{Toward Subgraph-Guided Knowledge Graph Question Generation with Graph Neural Networks}
%
%
%

\author{Yu~Chen,
        Lingfei~Wu,~\IEEEmembership{Member,~IEEE,}
        and~Mohammed~J.~Zaki,~\IEEEmembership{Fellow,~IEEE}
\thanks{Y. Chen is with Meta AI, Menlo Park,
CA 94025. Email: hugochen@meta.com.}
\thanks{L. Wu is with Pinterest, San Francisco, CA 94107. Email: teddy.lfwu@gmail.com.}%
\thanks{M. J. Zaki is with the Computer Science Department, Rensselaer Polytechnic Institute, Troy, NY 12180. Email: zaki@cs.rpi.edu.}
}

%
%

\markboth{Journal of \LaTeX\ Class Files,~Vol.~14, No.~8, August~2015}%
{Shell \MakeLowercase{\textit{et al.}}: Bare Demo of IEEEtran.cls for IEEE Journals}
%



\maketitle

\begin{abstract}
Knowledge graph (KG) question generation (QG) aims to generate natural language questions from KGs and target answers. Previous works mostly focus on a simple setting which is to generate questions from a single KG triple. In this work, we focus on a more realistic setting where we aim to generate questions from a KG subgraph and target answers. In addition, most of previous works built on either RNN-based or Transformer-based models to encode a linearized KG sugraph, which totally discards the explicit structure information of a KG subgraph. To address this issue, we propose to apply a bidirectional Graph2Seq model to encode the KG subgraph. Furthermore, we enhance our RNN decoder with node-level copying mechanism to allow directly copying node attributes from the KG subgraph to the output question. Both automatic and human evaluation results demonstrate that our model achieves new state-of-the-art scores, outperforming existing methods by a significant margin on two QG benchmarks. Experimental results also show that our QG model can consistently benefit the Question Answering (QA) task as a mean of data augmentation.
\end{abstract}


%
\IEEEpeerreviewmaketitle

\section{Introduction}
Recent years have seen a surge of interests in Question Generation (QG) in machine learning and natural language processing. The goal of QG is to generate a natural language (NL) question for a given form of data such as text \cite{du2017learning,song2018leveraging,chen2019reinforcement,pan2020semantic}, images \cite{li2018visual}, tables \cite{bao2018table}, and knowledge graphs (KGs) \cite{seyler2017knowledge}. In this work, we focus on QG from KGs.

One of the biggest applications of QG is to provide training data for question answering (QA) systems~\cite{tang2017question}.
KGs have drawn a large amount of research attention in recent years, partially due to their huge potential for an accessible, natural way of retrieving information without a need for learning complex query languages such as SPARQL. In order to train a large Knowledge Base Question Answering (KBQA) system~\cite{chen2019bidirectional,chen2021personalized}, a large number of labeled question-answer pairs are often needed, which can be a severe bottleneck in practice because human annotation is usually expensive and time-consuming.
Developing effective approaches to generate high-quality QA pairs can significantly address the data scarcity issue for KBQA.  
In addition, QG can be applied for educational purposes by producing practice assessments~\cite{heilman2010good}.
Moreover, QG can help dialog systems have more engaging conversations~\cite{mostafazadeh2016generating}.

\begin{figure}[!tb]
  \vspace{-3mm}
  \centering
    \includegraphics[keepaspectratio=true,scale=0.32]{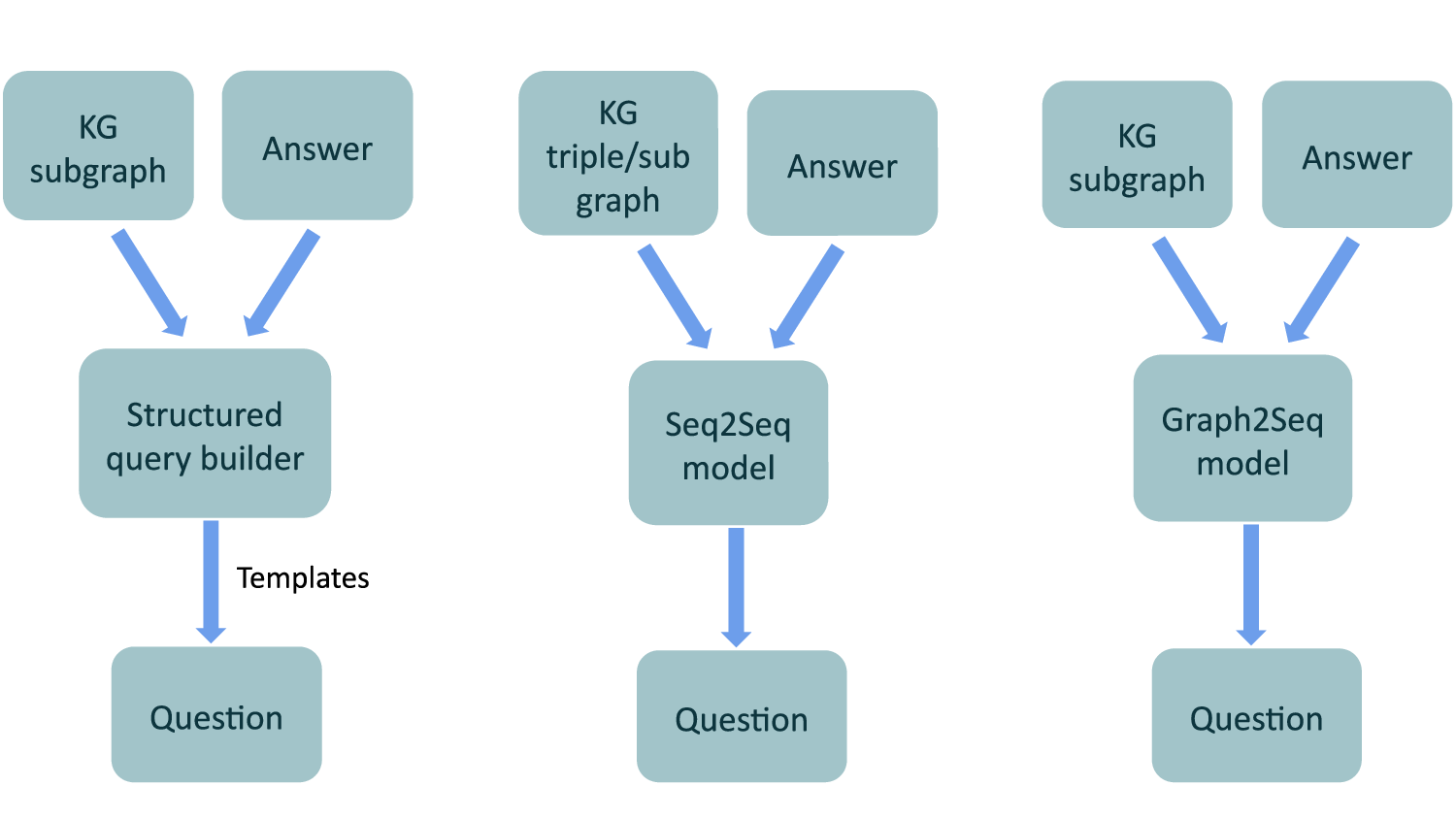}
  \caption{Various QG from KGs learning paradigms. Left: template-based approaches. Middle: Seq2Seq-based approaches for simple QG from a single KG triple or multi-hop QG from a KG subgraph. Right (ours): Graph2Seq-based approaches for multi-hop QG from a KG subgraph.}
  \label{fig:kgqg_types}
\vspace{-4mm}
\end{figure}

In the past decade, the research on QG from KGs has gained increasing interest and can be categorized into two classes. The first line of research heavily relies on handcrafted question templates~\cite{seyler2015generating,song2016question,seyler2017knowledge}. They typically first construct a structured query (e.g., SPARQL query) and then apply a template-based method to verbalize it to a natural language question. Using a set of pre-designed templates not only requires a significant amount of human effort, thus leading to low generalizability and scalability, but also limits the diversity and fluency of the generated questions. 
The other line of research adopts a purely data-driven end-to-end approach without resort to any handcrafted templates. Those are mostly neural network based approaches which employ an RNN or Transformer~\cite{vaswani2017attention} decoder to generate a natural language question as a sequence of tokens.
However, most of them~\cite{serban2016generating,reddy2017generating,elsahar2018zero,liu2019generating,hu2022generating} focus only on generating simple questions, which limits their usage in benefiting complex KBQA systems often requiring multi-hop reasoning. The main reason they can only generate simple questions is due to their incapability of encoding a KG subgraph containing a rich set of interlinked triples. Instead, they can only take a keyword list or a single KG triple (i.e., subject-predicate-object) as the input because they adopt a Sequence-to-Sequence (Seq2Seq)~\cite{sutskever2014sequence,cho2014learning} architecture which can only encode sequential data via a sequence encoder.

More recently, \cite{kumar2019difficulty} presented a Transformer-based Seq2Seq model named MHQG+AE for generating multi-hop complex questions from a KG subgraph. To the best of our knowledge, MHQG+AE was the first neural network-based model focusing on QG from a KG subgraph.
Because a Transformer cannot admit graph-structured input data like a KG subgraph, they proposed to represent a KG subgraph as a set of triples where the triple embeddings were computed based on the embeddings of the subject, predicate and object contained in the triple. They also removed the positional encoding in a regular Transformer in order to discard the position information of triples in a KG subgraph. Even though their approach was able to directly work on a KG subgraph for generating more complex questions compared to previous approaches, they failed to effectively utilize the rich structure information of a KG subgraph because they completely ignored the rich interactions among triples in a KG subgraph. 
In follow-up work, \cite{sheng2020knowledge} proposed to augment the input KG subgraph with external knowledge such as entity descriptions/domains, question word types and answer entity types.
However, they still failed to respect the rich structure information of the KG subgraph as they simply regarded a KG subgraph as a sequence of subject, predicate and object embeddings, and applied a bidirectional LSTM~\cite{hochreiter1997long} to learn its representations.
We believe capturing fine-grained structure information is critical for generating high-quality questions.

We summarize the three challenges of the task of multi-hop QG from KGs (denoted as KG-QG) as follows. The first one is how to learn a good representation of a KG subgraph. A KG subgraph has complex underlying structures such as node attributes and multi-relational edges. Each node and edge could have (long) associated text comprising multiple words. Previous approaches either only considered a keyword list or single triple for simple question generation or simply regarding the KG subgraph as a set of triples without fully utilizing its rich structure information when generating multi-hop questions.
The second challenge is how to automatically learn a good mapping between a subgraph and a natural language question.
For instance, it's common for a question to directly mention an entity name from the input KG subgraph. However, it's challenging for previous approaches to precisely generate such an entity name which often contains multiple tokens.
The third challenge is how to effectively leverage the answer information. Given a KG subgraph containing many triples, one can generate completely different questions without knowing the exact target answer. Therefore, effectively utilizing the answer information is crucial for generating more relevant questions.

In order to address the above challenges, we present a subgraph-guided knowledge graph question generation approach with Graph Neural Networks (GNNs). To this end, we introduce for the first time the Graph-to-Sequence (Graph2Seq) architecture with a novel node-level copying mechanism for the KG-QG task to address the second challenge. We extend the regular GNN-based encoder to allow processing directed and multi-relational KG subgraphs to solve the first challenge. 
In addition, we propose a simple yet elegant way to leverage the context information from the answers to effectively handle the third challenge. Extensive experimental results demonstrate that our model significantly outperforms the state-of-the-art baselines by a large margin on two benchmarks and consistently benefits the KBQA task. \cref{fig:kgqg_types} illustrates the main ideas of various QG from KGs learning paradigms.

We highlight our main contributions as follows:
\setlist{nolistsep}
\begin{itemize}[noitemsep]
    \item We propose a novel Graph2Seq model for subgraph-guided KG-QG. 
    The proposed Graph2Seq model employs bidirectional graph embedding and we design two different GNN encoders to effectively encode KG subgraphs with directed and multi-relational edges. 
    \item We extend the RNN decoder with a novel copying mechanism that allows the entire node attribute to be borrowed from the input KG subgraph when generating natural language questions.
    \item We investigate two different ways of initializing node/edge embeddings when applying a GNN encoder to process KG subgraphs. 
    In addition, we study the impact of edge direction on the GNN encoder.
    \item Experimental results show that our model improves the state-of-the-art BLEU-4 score from 11.57 to 29.40 and from 25.99 to 59.59 on WebQuestions (WQ) and PathQuestions (PQ) benchmarks, respectively.
    A human evaluation study corroborates that the questions generated by our model are more natural (semantically and syntactically) and relevant compared to other baselines. 
    Experiments also show that our QG model can consistently benefit the KBQA task as a mean of data augmentation.
\end{itemize}

\section{Related Work}

\subsection{Question Generation from Knowledge Graphs}\label{sec:related_work_nqg}

Early works~\cite{seyler2015generating,song2016question,seyler2017knowledge} on QG from KGs are mostly template-based approaches heavily relying on a set of pre-defined question templates to verbalize a structured query to a natural language question. However, they usually have low generalizability and scalability, and the diversity and fluency of the generated questions are limited due to the nature of template-based approaches.
Recently, Seq2Seq-based neural architectures have been applied to this task without resort to manually-designed templates and are end-to-end trainable.
However, these approaches~\cite{serban2016generating,reddy2017generating,elsahar2018zero,liu2019generating,hu2022generating} only focus on generating simple questions from a keyword list or single triple as they typically employ an RNN or Transformer based encoder which cannot handle graph-structured data like a KG subgraph.
Very recently, Seq2Seq-based approaches were also applied for generating a multi-hop complex question from a KG subgraph instead of just a single triple. However, they still failed to effectively utilize the rich structure information of the KG subgraph by simply regarding a KG subgraph as a set of triples~\cite{kumar2019difficulty} or a sequence of subject, predicate and object embeddings~\cite{sheng2020knowledge}.
Unlike all previous approaches, in this work, we focus on generating multi-hop complex questions by effectively modeling the rich structures (e.g., edge directions, edge types) of KG subgraphs via a novel GNN-based graph encoder. To the best of our knowledge, we are the first to introduce the Graph2Seq architecture to the KG-QG task.

There was related work focusing on QG from text.
In~\cite{chen2019reinforcement}, we proposed a Reinforcement Learning based Graph2Seq model for the task of QG from text. Besides the difference in terms of problem settings, the major technical difference between this work and our previous work includes, in this work, i) we extend the GNN encoder to handle multi-relational graphs where in~\cite{chen2019reinforcement} edge type information was not modeled, and ii) we extend the word-level copying mechanism in~\cite{chen2019reinforcement} to the node-level copying mechanism.
Some recent QG from text works explored leveraging the external knowledge for better performance. For instance, in~\cite{shen2022diversified}, the authors proposed to augment the raw text with auxiliary knowledge retrieved from a KG using entities and keywords mentioned in the input text. Their approach then applies three different GNN-based encoders to encode three types of graphs constructed based on text and knowledge retrieved from a KG. Even though we both adopt a Graph2Seq architecture, we tackle very different problems and they utilize KG as external knowledge for better QG from text performance.

Our work is also related to recent research efforts on pre-trained models for KG-to-text generation~\cite{chen2020kgpt,ke2021joint} which used KG-to-text generation as one of the pre-training tasks. These large-scale pre-trained models could be used for many downstream KG-to-text applications (including QG from KGs) by finetuning them for a particular downstream task.


\subsection{Graph Neural Networks}

Traditional Deep Learning approaches like Convolutional Neural Networks and Recurrent Neural Networks are designed for Euclidean data like images and text, and thus cannot directly handle non-Euclidean data like graphs. Over the past few years, Graph Neural Networks (GNNs)~\cite{kipf2016semi,gilmer2017neural,hamilton2017inductive,li2015gated,chen2020iterative,liu2022compact} have drawn increasing attention due to their ability to model graph-structured data and have successfully been applied in the NLP field~\cite{bastings2017graph,song2018graph,chen2020graphflow,wu2021graph,wu2021deep}.
Recently, in order to address the limitations of the widely used Seq2Seq architectures~\cite{sutskever2014sequence,cho2014learning} on encoding rich and complex graph-structured data, a number of works have applied the Graph2Seq architectures for various NLP tasks including machine translation~\cite{bastings2017graph,beck2018graph}, semantic parsing~\cite{xu2018exploiting}, code summarization~\cite{liu2021retrieval}, and graph-to-text generation (e.g., AMR, SQL and KG to text)~\cite{xu2018graph2seq,xu2018sql,marcheggiani2018deep,vougiouklis2018neural}.
Compared to existing Graph2Seq models, our proposed Graph2Seq model can better handle multi-relational graphs and employ node-level copying mechanism to enable generating more faithful text.

\section{Approach}

\begin{figure*}[!htb]
  \vspace{-3mm}
  \centering
    \includegraphics[keepaspectratio=true,scale=0.21]{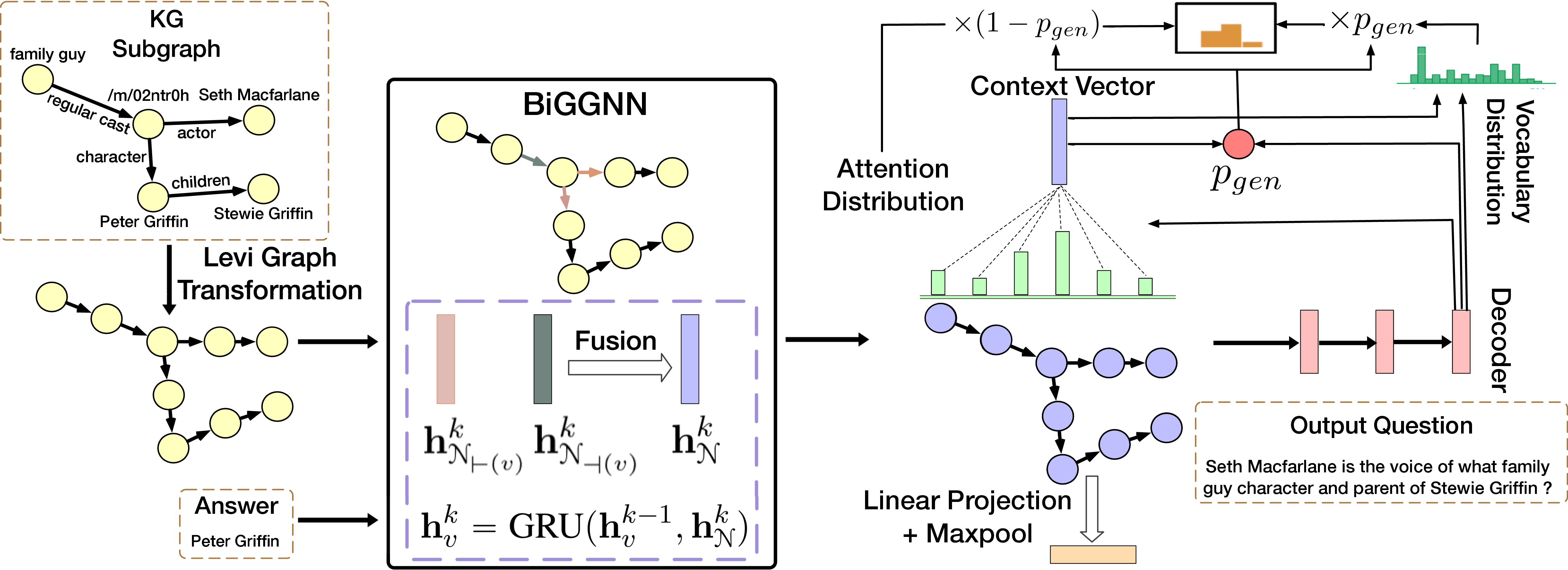}
  \caption{Overall architecture of our proposed model. Best viewed in color.}
  \label{fig:overall_arch}
\vspace{-3mm}
\end{figure*}

\subsection{Problem Formulation}

Our focus is on natural question generation from a KG subgraph, along with potential target answers; the overall architecture of our approach is shown in \cref{fig:overall_arch}. 
We assume that a KG subgraph is a collection of triples (i.e., subject-predicate-object), that can also be represented as a graph $\mathcal{G}=(V, E)$,
where $V \subseteq \mathcal{V}$ denotes a set of entities (i.e., subjects or objects) 
and $E \subseteq \mathcal{E}$ denotes all the predicates connecting these entities.
We denote by $\mathcal{V}$ and $\mathcal{E}$ the complete entity set and predicate set of the KG, respectively.
We also assume that all the answers from the target answer set $V^a$ are from the entity set $V$, which is the normal setting of the task of KBQA~\cite{chen2019bidirectional}.
The task of KG-QG is to generate the best natural language question consisting of a sequence of word tokens $\hat{Y}=\{y_1, y_2, ..., y_T\}$ which maximizes the conditional likelihood
$\hat{Y} = \argmax_Y P(Y|\mathcal{G}, V^a)$ where
$T$ is the length of the question.
We focus on the problem setting where we have a set of KG subgraphs (and answers) and target questions pairs, to learn the mapping; existing QG approaches \cite{serban2016generating,elsahar2018zero,kumar2019difficulty} make a similar assumption. 
Although the three main challenges we have discussed before are based on QG from KGs, other QG tasks from other data sources also share some or most of issues when dealing with these tasks. Therefore, our model could be generalized to cope with these tasks as well.

\subsection{Encoding Layer}

Let us denote $V$ as a set of nodes (i.e., entities) $\{v_1, v_2, ..., v_n\}$ in a KG subgraph $\mathcal{G}$, where each node is associated with some attributes such as text or ID.
Similarly, let us denote $E$ as a set of edges (i.e., predicates) $\{e_1, e_2, ..., e_m\}$ in $\mathcal{G}$, where each edge has some attributes such as text or ID.

\subsubsection{Encoding Nodes and Edges}

Before applying the GNN encoder to process a KG subgraph, 
we need to map nodes and edges to an initial embedding space that encodes their attributes.
There are two common ways of encoding nodes and edges in a KG.
One solution is based on global KG embeddings that are pretrained on the whole KG by some 
KG representation learning algorithm such as TransE~\cite{bordes2013translating},
while the other one is based on pretrained embeddings (e.g., GloVe~\cite{pennington2014glove}) of the words making up the textual attributes.
In this work, we choose to encode nodes and edges based on word embeddings of their textual attributes in our main model.
We posit that it is relatively easier for a model to learn the mapping from the input KG subgraph to the output NL question with both sides based on word embeddings.
We empirically compare and analyze the two encoding strategies in our experiments.
In order to encode the nodes and edges in a KG subgraph,
we apply two bidirectional LSTMs~\cite{hochreiter1997long} 
for nodes 
(i.e., one for nodes, and one for edges) to encode their associated text.
The concatenation of the last forward and backward hidden states of the BiLSTM is used as the initial embeddings for nodes and edges.

\subsubsection{Utilizing Target Answers}

In the setting of KBQA~\cite{chen2019bidirectional,haussmann2019foodkg}, it is usually assumed that the answers to a question are entities in a KG subgraph. 
As a dual task of KBQA, in this QG work, 
we assume that utilizing the target answers along with the KG subgraph can help generate more relevant questions.
To this end, we apply a simple yet effective strategy 
where we introduce an additional learnable markup vector associated with each node/edge to indicate whether it is an answer or not.
Therefore, the initial vector representation of a node/edge will be the concatenation of the BiLSTM output and the answer markup vector.
We denote $\vec{X}^e=\{\vec{x}^e_1, \vec{x}^e_2, ..., \vec{x}^e_n\}$ and $\vec{X}^p=\{\vec{x}^p_1, \vec{x}^p_2, ..., \vec{x}^p_m\}$ as the embeddings of the entity nodes and predicate edges, respectively.
Both $\vec{X}^e$ and $\vec{X}^p$ have the same embedding dimension $d$.

\subsection{Bidirectional Graph2Seq Generator with Copying}

While RNNs are good at modeling sequential data, they cannot naturally handle graph-structured data.
One might need to linearize a graph to a sequence so as to apply an RNN-based encoder, which will lose the rich structure information in the graph.
Many previous works~\cite{bastings2017graph,beck2018graph} showed the superiority of GNNs compared to RNNs on modeling graph-structured data. 
\cite{kumar2019difficulty} proposed to encode a set of triples via a Transformer by removing positional encoding in the original architecture.
Even though a Transformer-based encoder could learn the semantic relations among the triples through the all-to-all attention,
the explicit graph structure is totally discarded.
In this work, we introduce a 
bidirectional GNN-based encoder to encode the KG subgraph, and decode the output question via an RNN-based decoder equipped with node-level copying mechanism.


\subsubsection{Bidirectional Graph Encoder}

Many existing GNNs~\cite{kipf2016semi,hamilton2017inductive,velivckovic2017graph} were not designed to process directed graphs such as a KG.
Even though some GNN variants such as GGSNN~\cite{li2015gated} and MPNN~\cite{gilmer2017neural} are able to handle directed graphs via message passing across graphs, 
they do not model the bidirectional information when aggregating information from neighboring nodes for each node.
As a result, messages can only be passed across graphs in a unidirectional way.

In this work, we introduce the Bidirectional Gated Graph Neural Network (BiGGNN) which extends GGSNN
by learning node embeddings from both incoming and outgoing directions in an interleaved fashion when processing a directed graph.
A similar bidirectional approach has been exploited in \cite{xu2018graph2seq,ribeiro2019enhancing} to extend other GNN variants. 
While their methods simply learn the node embeddings of each direction independently and concatenate them at last step,
BiGGNN fuses the intermediate node embeddings from both directions at every iteration.

The embedding $\vec{h}_v^{0}$ for node $v$ is initialized to $\vec{x}_v$, namely, a concatenation of the BiLSTM output and the answer markup vector. 
BiGGNN then performs message passing across the graph for a fixed number of hops, with the same set of network parameters shared at each hop.
At each hop of computation, for every node in the graph, we apply an aggregation function that takes as input a set of incoming (or outgoing) neighboring node vectors and outputs a backward (or forward) aggregation vector.
In principle, many order-invariant operators such as max or attention \cite{velivckovic2017graph} can be employed to aggregate neighborhood information.
Here we use a simple average aggregator:
\begin{equation}\label{eq:node_agg}
\begin{aligned}
\vec{h}^k_{\EuScript{N}_{\dashv(v)}}=\text{AVG}(\{\vec{h}^{k-1}_v\} \cup \{\vec{h}^{k-1}_u, \forall u \in \EuScript{N}_{\dashv(v)}\})\\
\vec{h}^k_{\EuScript{N}_{\vdash(v)}}=\text{AVG}(\{\vec{h}^{k-1}_v\} \cup \{\vec{h}^{k-1}_u, \forall u \in \EuScript{N}_{\vdash(v)}\})
\end{aligned}
\end{equation}
where $\EuScript{N}_{\dashv(v)}$ and $\EuScript{N}_{\vdash(v)}$ denote the incoming and outgoing neighbors of node $v$. 
We then fuse the node embeddings aggregated from both directions,
\begin{equation}
\begin{aligned}
\vec{h}^k_{\EuScript{N}_{(v)}}=\text{Fuse}(\vec{h}^k_{\EuScript{N}_{\dashv(v)}}, \vec{h}^k_{\EuScript{N}_{\vdash(v)}})  
\end{aligned}
\end{equation}

The fusion function is computed as a gated sum of two information sources,
\begin{equation}
\begin{aligned}
\text{Fuse}(\vec{a}, \vec{b}) = \vec{z} \odot \vec{a} + (1-\vec{z}) \odot \vec{b}\\
\vec{z} = \sigma(\vec{W}_{\!z} [\vec{a}; \vec{b}; \vec{a} \odot \vec{b}; \vec{a}-\vec{b}]+\vec{b}_z)
\end{aligned}
\end{equation}
where $\odot$ is the component-wise multiplication, $\sigma$ is a sigmoid function, and $\vec{z}$ is a gating vector.
The gate helps the model to determine how much of the information needs to be reserved from the two aggregated node embeddings.

Finally, a Gated Recurrent Unit (GRU)~\cite{cho2014learning} is used to update the node embeddings by incorporating the aggregation information.
\begin{equation}
\begin{aligned}
\vec{h}^{k}_v=\text{GRU}(\vec{h}^{k-1}_v, \vec{h}^k_{\EuScript{N}_{(v)}})
\end{aligned}
\end{equation}
After $n$ hops of GNN computation where $n$ is a hyperparameter,
we obtain the final state embedding $\vec{h}^{n}_v$ for node $v$.
To compute the graph-level embedding, we first apply a linear projection to the node embeddings, and then apply max-pooling over all node embeddings to get a $d$-dim vector $\vec{h}^\mathcal{G}$.

\subsubsection{Handling Multi-relational Graphs}

KGs are typically heterogeneous networks that contain a large number of edge types.
However, many existing GNNs~\cite{kipf2016semi,hamilton2017inductive,li2015gated,velivckovic2017graph} are not directly applicable to multi-relational graphs.
In order to model both node and edge information with GNNs,
researchers have extended them by either having separate learnable weights for different edge types or having explicit edge embeddings when performing message passing~\cite{gilmer2017neural,simonovsky2017dynamic}.
While the former solution may have severe scalability issues when handling graphs with a large number of edge types, 
the later one requires major modifications to existing GNNs.
In this work, we explore two solutions to adapt GNNs to multi-relational graphs.

\smallskip
\noindent\textbf{Levi graph transformation.} 
We can directly apply regular GNNs to a multi-relational KG subgraph
by converting it to a Levi graph~\cite{levi1942finite}.
Specifically, we treat all edges in the original graph as new nodes and add new edges connecting original nodes and new nodes, which results in a bipartite graph.
For instance, in a KG subgraph, a triple (Mario\_Siciliano, place\_of\_birth, Rome)
will be converted to ``Mario\_Siciliano $\rightarrow$ place\_of\_birth $\rightarrow$ Rome'' where ``place\_of\_birth'' becomes a new node, and $\rightarrow$ indicates a new edge connecting an entity and a predicate.
Note that since most KG subgraphs are sparse, the number of newly added nodes (and edges as well) will at most be linear to the number of original nodes.

\smallskip
\noindent\textbf{Gated message passing with edge information.}
We also extend BiGGNN to explicitly incorporate edge embeddings when conducting message passing, calling the resultant variant as BiGGNN$_{edge}$.
Specifically, we rewrite the node aggregation function~\cref{eq:node_agg}
as follows,
\begin{equation}\label{eq:node_edge_agg}
\small
\begin{aligned}
\vec{h}^k_{\EuScript{N}_{\dashv(v)}}=\text{AVG}(\{\vec{h}^{k-1}_v\} \cup \{f([\vec{h}^{k-1}_u; \vec{e}_{uv}]), \forall u \in \EuScript{N}_{\dashv(v)}\})\\
\vec{h}^k_{\EuScript{N}_{\vdash(v)}}=\text{AVG}(\{\vec{h}^{k-1}_v\} \cup \{f([\vec{h}^{k-1}_u; \vec{e}_{uv}], \forall u \in \EuScript{N}_{\vdash(v)}\})
\end{aligned}
\end{equation}
where f is a nonlinear function (i.e., linear projection + ReLU~\cite{nair2010rectified}) applied to the concatenation of $\vec{h}^{k-1}_u$ and $\vec{e}_{uv}$ which is
the embedding of the edge connecting node $u$ and $v$.

\subsubsection{RNN Decoder with Node-level Copying}

We adopt an attention-based~\cite{bahdanau2014neural,luong2015effective} LSTM decoder that generates the output sequence one word at a time.
The decoder takes the graph-level embedding $\vec{h}^\mathcal{G}$ followed by two separate fully-connected layers as initial hidden states (i.e., $\vec{c}_0$ and $\vec{s}_0$) and the node embeddings $\{\vec{h}^{n}_v, \forall v \in \mathcal{G}\}$ as the attention memory.
The particular attention mechanism used in our decoder closely follows~\cite{see2017get}.
Basically, at each decoding step $t$, an attention mechanism learns to attend to the most relevant nodes in the input graph, and computes a context vector $\vec{h}^*_t$ based on the current decoding state $\vec{s}_t$ and the attention memory.

We hypothesize that when generating NL questions from a KG subgraph, it is very likely to directly mention (i.e., copy) entity names that are from the input KG subgraph even without rephrasing them.
When augmented with copying mechanism~\cite{vinyals2015pointer,gu2016incorporating}, most RNN decoders are typically allowed to copy words from the input sequence.
We extend the regular word-level copying mechanism to the node-level copying mechanism that
allows copying node attributes (i.e., node text) from the input graph.
Copying mechanism was used in some previous Graph2Seq papers~\cite{marcheggiani2018deep,koncel2019text}. The most similar work is~\cite{koncel2019text} which proposed to copy both entities and predicates from the input graph. Unlike~\cite{koncel2019text}, we use masked copying mechanism to only copy entity nodes in the transformed Levy graph and do not copy predicate nodes. This is because we assume that for the KG-QG task, it is very likely for humans to directly mention entity names but not necessarily for predicate names that are from the KG.

At each decoding step,
the generation probability $p_\text{gen} \in [0, 1]$ is calculated from the context vector $\vec{h}^*_t$, the decoder state $\vec{s}_t$ and the decoder input $y_{t-1}$.
Next, $p_\text{gen}$ is used as a soft switch to choose between generating a word from the vocabulary or copying a node attribute from the input graph.
We dynamically maintain an extended vocabulary which is the union of the usual vocabulary and all node names appearing in a batch of source examples (i.e., KG subgraphs).

\subsection{Training and Testing}
As customary for training sequential models,
we minimize the following cross-entropy loss,
\vspace{-1mm}
\begin{equation}~\label{eq:cross_entropy_loss}
\mathcal{L} = \sum_{t}{}-\log{} P(y_t^*|X, y^*_{<t})
\end{equation}
\vspace{-1mm}
where $y_t^*$ is the word at the $t$-th position of the gold output sequence.
Scheduled teacher forcing~\cite{bengio2015scheduled} is adopted to alleviate the exposure bias problem.
During the testing phase, beam search is applied to generate the output.

\noindent\textbf{Two-stage training strategy}.

Most prior works on QG employ cross-entropy based training objective, which is also a de facto choice for training sequential models in many other NLP tasks.
However, cross-entropy based training strategy has some known limitations including exposure bias and evaluation discrepancy between training and testing~\cite{ranzato2015sequence,wu2016google,paulus2017deep}.
That is to say, during training, a model has access to the ground-truth previous token when decoding and is optimized toward cross-entropy loss, while during testing, 
no ground-truth previous token is provided and cross-entropy loss is not used for evaluation.

To tackle these issues, besides training our proposed model with the regular cross-entropy loss, we also explore a two-stage training strategy where we first train the model with cross-entropy loss, and then finetune the model with a hybrid loss combining both the cross-entropy loss and Reinforcement Learning (RL)~\cite{williams1992simple} loss.
The RL loss is defined based on evaluation metrics, enabling us to directly optimize the model towards the evaluation metrics.

The reason we need the first stage training is because training models from scratch using RL is often challenging. The regular cross-entropy training can help us obtain a reasonably good performing model, and the RL-based finetuning can further improve the model performance.

In the first stage, the regular cross-entropy loss is used,
\begin{equation}
\mathcal{L}_{lm} = \sum_{t}{}-\log{} P(y_t^*|X, y^*_{<t})
\end{equation}
as in~\cref{eq:cross_entropy_loss}. 
In the second stage, we further finetune the model by optimizing a hybrid objective function combining both cross-entropy loss and RL loss, defined as,
\begin{equation}\label{eq:mixed_loss}
\begin{aligned}
\mathcal{L}= \gamma \mathcal{L}_{rl} + (1 - \gamma) \mathcal{L}_{lm}
\end{aligned}
\end{equation}
where $\gamma$ is a scaling factor controlling the trade-off between the two losses.

While our architecture is agnostic to the specific RL algorithm, in this work, 
we employ an efficient yet effective RL approach called self-critical sequence training (SCST)~\cite{rennie2017self}
to directly optimize the discrete evaluation metrics. 
SCST is an efficient REINFORCE algorithm that utilizes the output of its own test-time inference algorithm to normalize the rewards it experiences.
At each training iteration,
the RL loss is defined by comparing the reward of the sampled output $Y^s$
with the reward of the baseline output $\hat{Y}$, 
\begin{equation}
\mathcal{L}_{rl}=(r(\hat{Y})-r(Y^s)) \sum_{t}{\log{} P(y_t^s|X, y^s_{<t})}
\end{equation}
where $Y^s$ is produced by multinomial sampling, that is, each word $y_t^s$ is sampled according to the likelihood $P(y_t|X, y_{<t})$ predicted by the generator,
and $\hat{Y}$ is obtained by greedy search, that is, by maximizing the output probability distribution at each decoding step.
As we can see,  minimizing the above loss is equivalent to maximizing the likelihood of some sampled output that has a higher reward than the corresponding baseline.

One of the key factors for RL is to pick the proper reward function.
We define $r(Y)$ as the reward of an output sequence $Y$, computed by comparing it to the corresponding ground-truth sequence $Y^*$ with some reward metric which is 
a combination of our evaluation metrics (i.e., we used BLEU-4 and ROUGE-L scores in our experiments).
This lets us directly optimize the model towards the evaluation metrics.

\section{Experiments}

In this section, we conduct extensive experiments to evaluate the effectiveness of our proposed model for the QG task.
We also conduct experiments to examine whether our QG model can help the QA task by providing more training data.
Besides, we want to examine whether the introduced GNN-based encoder works better than an RNN-based or Transformer-based encoder when encoding a KG subgraph for the QG task.
In addition, we explore and analyze two different ways of handling multi-relational graphs with GNNs.
Moreover, we empirically compare two different ways of initializing node and edge embeddings before feeding them into a GNN-based encoder.
An experimental comparison between bidirectional GNN-based encoder and unidirectional GNN-based encoder is also provided.
The code and data will be released upon the paper acceptance.

\subsection{Baseline Methods}

We compare our model against the following baselines:
i) L2A~\cite{du2017learning},
ii) Transformer (w/ copy)~\cite{vaswani2017attention},
iii) MHQG+AE~\cite{kumar2019difficulty},
iv) JointGT (T5)~\cite{ke2021joint},
and v) JointGT (BART)~\cite{ke2021joint}.
To the best of our knowledge, MHQG+AE was probably the first neural network-based model that focused on QG from a KG subgraph.
Their proposed model, called MHQG+AE, employs a Transformer-based encoder~\cite{vaswani2017attention} to encode a KG subgraph (i.e., a set of triples), and generates an output question with a Transformer-based decoder.
L2A is a LSTM-based Seq2Seq model equipped with attention mechanism, which takes as input a linearized KG subgraph.
It was included in \cite{kumar2019difficulty} as a baseline.
The results of L2A reported here are taken from~\cite{kumar2019difficulty}.
We also include a Transformer-based encoder-decoder model~\cite{klein2017opennmt} with copying mechanism that takes as input a linearized KG subgraph, i.e., a sequence of triples where each triple is represented as a sequence of tokens containing the subject name, predicate name and object name.
Hence, after the transformation, a KG subgraph becomes a sequence of tokens.
Note that the Transformer baseline included in our experiments encodes the word sequence that is linearized from a KG subgraph,
while the MHQG+AE model encodes the triple set contained in a KG subgraph by removing the positional encoding in a regular Transformer architecture.
Unlike MHQG+AE that takes as input a set of triple embeddings that are pretrained by a knowledge-base representation learning framework called TransE~\cite{bordes2013translating},
the Transformer baseline takes a sequence of word embeddings as input.
We used the open-source implementation~\cite{klein2017opennmt} of the Transformer-based encoder-decoder model that is equipped with copying mechanism.
Lastly, we also include two large-scale pre-trained KG-to-text models JointGT (T5) and JointGT (BART)~\cite{ke2021joint} which were finetuned for the task of QG from KGs. We do not include~\cite{sheng2020knowledge} as our baseline because their approach augmented the input KG subgraph with various types of external knowledge such as entity descriptions, entity domains, question word types and answer entity types, which makes it unfair to directly compare the performance of their approach with our approach. In their original paper~\cite{sheng2020knowledge}, the authors reported that their ablated system without using auxiliary knowledge (i.e., but it still utilized the additional question word type information, see the results in their Table 3) significantly underperformed our approach (denoted as BiGraph2Seq in their Table 2) on two benchmarks (i.e., 3.11 absolute BLEU-4 gap and 0.64 absolute BLEU-4 gap).

\subsection{Data and Metrics}

\begin{table*}[!htb]

\captionsetup{font=normalsize}
\caption{Data statistics. The min/max/avg statistics are reported on the queries and KG subgraph triples.}
\label{table:data_stat}
\vspace{-3mm}
\begin{center}
\scalebox{1}{
\begin{tabular}{llllll}
\hline
Data &  \# examples & \# entities &  \# predicates &  \# triples  & query length \\
  \hline
WQ & \quad 22,989 &  \quad 25,703 & \quad\quad 672 & 2/99/5.8  & \quad 5/36/15\\
PQ & \quad 9,731 & \quad 7,250 & \quad\quad 378 & 2/3/2.7 &  \quad 8/25/14\\
 \hline
\end{tabular}
}
\end{center}
\vspace{-4mm}
\end{table*}

Following~\cite{kumar2019difficulty}, we used WebQuestions (WQ) and PathQuestions (PQ)
\footnote{\url{https://github.com/liyuanfang/mhqg}}
as our benchmarks where both of them use Freebase~\cite{freebase:datadumps} as the underlying KG.
The WQ dataset combines examples from WebQuestionsSP~\cite{yih2016value} and  ComplexWebQuestions~\cite{talmor2018web} where both of them are KBQA benchmarks that contain natural language questions, corresponding SPARQL queries and answer entities.
For each instance in WQ, in order to construct the KG subgraph, ~\cite{kumar2019difficulty} converted its SPARQL query to return a subgraph instead of the answer entity, by changing it from a SELECT query to a CONSTRUCT query. 
The WQ dataset~\cite{kumar2019difficulty} contains 18,989/2,000/2,000 (train/development/test) examples.
The PQ dataset~\cite{zhou2018interpretable} is similar to WQ except that
the KG subgraph in PQ is a path between two entities that span two or three hops.
The PQ dataset contains 9,793/1,000/1,000 (train/development/test) examples.
Brief statistics of the two datasets are provided in~\cref{table:data_stat}.

Following previous works, we use BLEU-4~\cite{papineni2002bleu}, METEOR~\cite{banerjee2005meteor} and ROUGE-L~\cite{lin2004rouge}
as automatic evaluation metrics. 
Initially, BLEU-4 and METEOR were designed for evaluating machine translation systems and ROUGE-L was designed for evaluating text summarization systems.
We also conduct a human evaluation study on WQ.
Generated questions are rated (i.e., range 1-5) based on whether they are syntactically correct, semantically correct and relevant to the KG subgraph.
More specially, we conducted a small-scale (i.e., 50 random examples per system) human evaluation study on the WQ test set.
We asked 6 human evaluators to give feedback on the quality of questions generated by a set of anonymized competing systems.
In each example, given a KG subgraph, target answers and an anonymized system output, they were asked to rate the quality of the output by answering the following three questions: i) is this generated question syntactically correct? ii) is this generated question semantically correct? and iii) is this generated question relevant to the KG subgraph and target answers?
For each evaluation question, the rating scale is from 1 to 5 where a higher score means better quality (i.e., 1: Poor, 2: Marginal, 3: Acceptable, 4: Good, 5: Excellent).
Responses from all evaluators were collected and averaged.

\subsection{Model Settings}\label{sec:model_settings}

We keep and fix the 300-dim GloVe~\cite{pennington2014glove} vectors for those words that occur more than twice in the training set.
The dimensions of answer markup embeddings are set to 32 and 24 for WQ and PQ, respectively.
We set the hidden state size of BiLSTM to 150 so that the concatenated state size for both directions is 300.
The size of all other hidden layers is set to 300.
We apply a variational dropout~\cite{kingma2015variational} rate of 0.4 after word embedding layers and 0.3 after RNN layers.
The label smoothing ratio is set to 0.2.
The number of GNN hops is set to 4.
During training, 
in each epoch, we set the initial teacher forcing probability to 0.8 and exponentially increase it to $0.8 * 0.9999^i$ where $i$ is the training step.
In addition, partial teacher forcing is adopted, which means that
when generating a sequence, some steps can be teacher forced and some not.
We use Adam \cite{kingma2014adam} as the optimizer.
The learning rate is set to 0.001.
We reduce the learning rate by a factor of 0.5 if the validation BLEU-4 score stops improving for three epochs. 
We stop the training when no improvement is seen for 10 epochs.
We clip the gradient at length 10.
The batch size is set to 30.
The beam search width is set to 5.
In the RL fine-tuning experiments, 
we set $\gamma$ in the mixed loss function~\cref{eq:mixed_loss} to 0.02 for WQ and 0.07 for PQ.
And the ratios of BLEU-4 score and ROUGE-L score for computing the reward are set to 1 and 0.02, respectively.
We set the learning rate to 0.00001 and 0.00002 for WQ and PQ, respectively.
All hyperparameters are tuned on the development set.
Experiments were conducted on a machine which has an Intel i7-2700K CPU and an Nvidia Titan Xp GPU with 16GB RAM.

\subsection{Experimental Results}

\begin{table*}[tb]
\vspace{-3mm}
\caption{Automatic evaluation results on WQ and PQ. The methods marked with $^{\dagger}$ are large-scale pre-trained KG-to-text models finetuned on QG data while other methods do not have access to such pre-training data. The results marked in bold and with $^{*}$ indicate the best and second best results, respectively.}
\label{table:exp_results}
\begin{center}
\scalebox{1}{
\begin{tabular}{lllllllll}
\hline
 \multirow{2}{*}{Method}& \vline &  &\quad WQ&& \vline & &\quad PQ& \\
    & \vline &  BLEU-4 &   METEOR &  ROUGE-L &  \vline &   BLEU-4 &   METEOR &  ROUGE-L\\
  \hline
 L2A & \vline & \quad6.01 &\quad 25.24 &\quad 26.95 & \vline & \quad17.00 &\quad 19.72 &\quad 50.38\\
 Transformer & \vline & \quad 8.94 & \quad13.79 & \quad32.63 & \vline & \quad56.43 &\quad 43.45 & \quad73.64\\
 MHQG+AE & \vline & \quad11.57 & \quad29.69 &\quad 35.53 & \vline & \quad25.99 & \quad33.16 & \quad58.94\\
  \hline
  JointGT (T5)$^{\dagger}$ & \vline & \quad 28.95& \quad 31.29$^{*}$&\quad 54.47 & \vline & \quad 60.45& \quad 45.38$^{*}$ & \quad77.59\\
  JointGT (BART)$^{\dagger}$ & \vline & \quad \textbf{30.02} & \quad \textbf{32.05}&\quad  \textbf{55.60}& \vline & \quad \textbf{65.89}& \quad \textbf{48.25} & \quad \textbf{78.87}\\
  \hline
 G2S+AE (ours) &  \vline & \quad 29.45$^{*}$ & \quad30.96 & \quad 55.45$^{*}$ & \vline & \quad 61.48$^{*}$ & \quad44.57 & \quad 77.72$^{*}$\\
 G2S$_{edge}$ +AE (ours) &  \vline &  \quad29.40 &\quad 31.12  & \quad55.23 & \vline &  \quad 59.59  & \quad 44.70 &  \quad75.20\\
 \hline
\end{tabular}
} 
\end{center}
\vspace{-2mm}
\end{table*}

\begin{table*}[!htb]
\vspace{-1mm}
\caption{Human evaluation results ($\pm$ standard deviation) on the WQ test set.
The rating scale is from 1 to 5 
(higher scores indicate better results).}
\label{table:wq_human_evaluation_results}
\begin{center}
\scalebox{1}{
\begin{tabular}{llllll}
\hline
Method & \vline & Syntactic  & Semantic  & Relevant & Overall\\
  \hline
  \text{Transformer} & \vline & \textbf{4.53 (0.18)} & \textbf{4.58 (0.22)} & 2.65 (0.57) & 3.92 (0.24)\\
 \text{G2S+AE} & \vline &4.18 (0.30) & 4.30 (0.27)& 4.26 (0.34) &4.25 (0.26) \\
  Ground-truth & \vline & 4.30 (0.15) &  4.50 (0.18) & \textbf{4.32 (0.32)} & \textbf{4.38 (0.19)}\\
 \hline
\end{tabular}
}
\end{center}

\vspace{-3mm}
\end{table*}

\begin{table*}[!htb]
\vspace{-1mm}
\caption{Ablation study on WQ and PQ.}
\label{table:ablation_results}
\begin{center}
\scalebox{1}{
\begin{tabular}{lllllllll}
\hline
 \multirow{2}{*}{Method}& \vline &  &\quad WQ&& \vline & &\quad PQ& \\
    & \vline &  BLEU-4 &  METEOR &   ROUGE-L &  \vline &  BLEU-4 &   METEOR &  ROUGE-L\\
  \hline
G2S+AE &  \vline & \quad\textbf{29.45} & \quad\textbf{30.96} & \quad\textbf{55.45} & \vline & \quad\textbf{61.48} & \quad\textbf{44.57} &\quad\textbf{77.72}\\
G2S & \vline & \quad28.43 & \quad30.13 & \quad54.44 & \vline & \quad60.68 &\quad 44.07 & \quad75.94\\
G2S w/o copy & \vline &\quad 22.95 & \quad26.99 & \quad51.05 & \vline &\quad 57.10 & \quad42.66 &\quad 74.29\\

 \hline
\end{tabular}
}
\end{center}
\vspace{-4mm}
\end{table*}

\noindent\textbf{Automatic evaluation results.}
\cref{table:exp_results} shows the evaluation results comparing our proposed models against other state-of-the-art baseline methods on WQ and PQ test sets.
As we can see, our models outperform all QG baselines by a large margin on both benchmarks.
This verifies the effectiveness of the proposed model.
Besides, we can clearly see the advantages of GNN-based encoders for modeling KG subgraphs, by comparing our model with RNN-based (i.e., L2A) and Transformer-based (i.e., Transformer, MHQG+AE) baselines.
Compared to our Graph2Seq model, both RNN-based and Transformer-based baselines ignore the explicit graph structure of a KG subgraph, which leads to degraded performance.
Although RNNs are suitable for processing sequential data such as text, they are incapable of modeling graph-structured data such as a KG subgraph.
To apply the RNN-based L2A model to a KG subgraph, 
\cite{kumar2019difficulty} linearized the graph to a sequence during preprocessing.
However, this inevitably ignores the rich structure information in the graph.
Recently, the Transformer~\cite{vaswani2017attention} has become a good alternative to the RNN when processing sequential data.
Even though a Transformer might be able to learn the semantic relations among the sequence elements through all-to-all attention, 
the explicit graph structure of a KG subgraph is totally discarded by the model.
Given these limitations, as shown in our experiments,  both of the two Transformer-based Seq2Seq baselines significantly underperform our GNN-based Graph2Seq model.
Interestingly, the Transformer baseline performs reasonably well on PQ,
but dramatically fails on WQ.
We speculate this is because PQ is more friendly to sequential models such as Transformer as the KG subgraph in PQ is more like path-structure
while the one in WQ is more like tree-structure.

The comparisons with large-scale pre-trained KG-to-text models further demonstrated the superiority of our models. Without access to a large amount of pre-training data, our best performing model clearly outperforms the large-scale model JointGT (T5) and achieves competitive results compared to JointGT (BART).
We also compare two variants of our model (i.e., G2S vs. G2S$_{edge}$) for handling multi-relational graphs.
As shown in ~\cref{table:exp_results}, directly applying the BiGGNN encoder to a Levi graph converted from a KG subgraph works quite well.
The proposed BiGGNN$_{edge}$ model can directly handle multi-relational graphs without modifying the input graph.
However, it performs slightly worse than the Levi graph solution.
Future directions of improving BiGGNN$_{edge}$ include updating edge embeddings in the message passing process and attending to edges in the attention mechanism.

\noindent\textbf{Human evaluation results.}
We conduct a human evaluation study to assess the quality of the questions generated by our model, the Transformer baseline, and the ground-truth data in terms of syntax, semantics and relevance metrics. 
In addition, an overall score is computed for each example by taking the average of the three scores.
As shown in~\cref{table:wq_human_evaluation_results},
overall, we can see that our model achieves good results even compared to the ground-truth, and outperforms the Transformer baseline.
Interestingly, we observe that the Transformer baseline gets high syntactic and semantic scores, but very poor relevant scores.
After manually examining some generated questions,
we noticed that it generates many fluent and meaningful questions that are by no means relevant to the given KG subgraph.
However, our model is able to generate more relevant questions possibly by better capturing the KG semantics and the answer.

\subsection{Ablation Study}
As shown in~\cref{table:ablation_results}, we perform an ablation study to assess the performance impacts of different model components.
First of all, the node-level copying mechanism contributes a lot to the overall model performance.
By turning it off, we observe significant performance drops on both benchmarks.
This verifies our assumption that when generating questions from a KG subgraph,
one usually directly copies named entities from the input KG subgraph to the output question.
Besides, the answer information is also important for generating relevant questions.
Even with the simple answer markup technique, we can see the performance boost on both benchmarks.

\subsection{Model Analysis}

\subsubsection{Effect of Node/Edge Embedding Initialization}\label{sec:node_edge_emb}
We empirically compare two different ways of initializing node/edge embeddings when applying the Graph2Seq model.
As shown in~\cref{table:kg_init_results},  encoding nodes and edges based on word embeddings of their textual attributes works better than based on their KG embeddings.
This might be because it is difficult for a NN-based model to learn the gap between KG embeddings on the encoder side and word embeddings on the decoder side.
 With the word embedding-based encoding strategy, it is relatively easier for a model to learn the mapping from the input KG subgraph to the output NL question.
 It also seems that modeling local dependency within the  subgraph without utilizing the global KG information 
 is enough for generating meaningful questions from a KG subgraph.

\begin{table}[htb]
\vspace{-2mm}
\caption{Effect of node/edge init. embeddings on WQ.}
\label{table:kg_init_results}
\begin{center}
\addtolength{\tabcolsep}{-3.8pt}
\scalebox{1}{
\begin{tabular}{lllll}
\hline
Method& \vline &   BLEU-4 &   METEOR &   ROUGE-L \\
  \hline
w/ word emb. & \vline & \quad28.43 & \quad30.13 & \quad54.44 \\
w/ KG emb. & \vline & \quad22.80 & \quad25.85 & \quad48.93 \\
 \hline
\end{tabular}
}
\end{center}
\vspace{-3mm}
\end{table}

\begin{table}[tbh]
\vspace{-3mm}
\caption{Impact of directionality for G2S+AE on PQ.}
\label{table:directionality_gnn_results}
\begin{center}
\addtolength{\tabcolsep}{-3.5pt}
\scalebox{1}{
\begin{tabular}{lllll}
\hline
Method& \vline &   BLEU-4 &   METEOR &   ROUGE-L \\
  \hline
Bidirectional & \vline & \quad 61.48 & \quad44.57 &  \quad77.72\\
Forward & \vline &  \quad59.59 & \quad42.72 & \quad75.82\\
Backward & \vline &   \quad59.12 &\quad 42.66 & \quad75.03\\
 \hline
\end{tabular}
}
\end{center}
\vspace{-3mm}
\end{table}

\begin{table}[!htb]
\vspace{-3mm}
\caption{Results of RL-based G2S+AE on WQ.}
\label{table:wq_rl_results}
\begin{center}
\addtolength{\tabcolsep}{-3.8pt}
\scalebox{1}{
\begin{tabular}{lllll}
\hline
Method& \vline &   BLEU-4 &   METEOR &   ROUGE-L \\
  \hline
G2S+AE & \vline & \quad 29.45 & \quad30.96 & \quad55.45 \\
G2S+AE+RL & \vline & \quad 29.80 &\quad 31.29  & \quad 55.51 \\
 \hline
\end{tabular}
}
\end{center}
\vspace{-3mm}

\end{table}

\begin{table}[!htb]
\vspace{-3mm}
\caption{Results of RL-based G2S+AE on PQ.}
\label{table:pq_rl_results}
\begin{center}
\addtolength{\tabcolsep}{-3.8pt}
\scalebox{1}{
\begin{tabular}{lllll}
\hline
Method& \vline &   BLEU-4 &   METEOR &   ROUGE-L \\
  \hline
G2S+AE & \vline &\quad61.48 & \quad44.57 & \quad77.72\\
G2S+AE+RL & \vline & \quad 59.21 &\quad44.47  &\quad77.35 \\
 \hline
\end{tabular}
}
\end{center}
\vspace{-2mm}

\end{table}

\begin{figure}[!htb]
  \vspace{-1mm}
  \centering
   \includegraphics[keepaspectratio=true,scale=0.15]{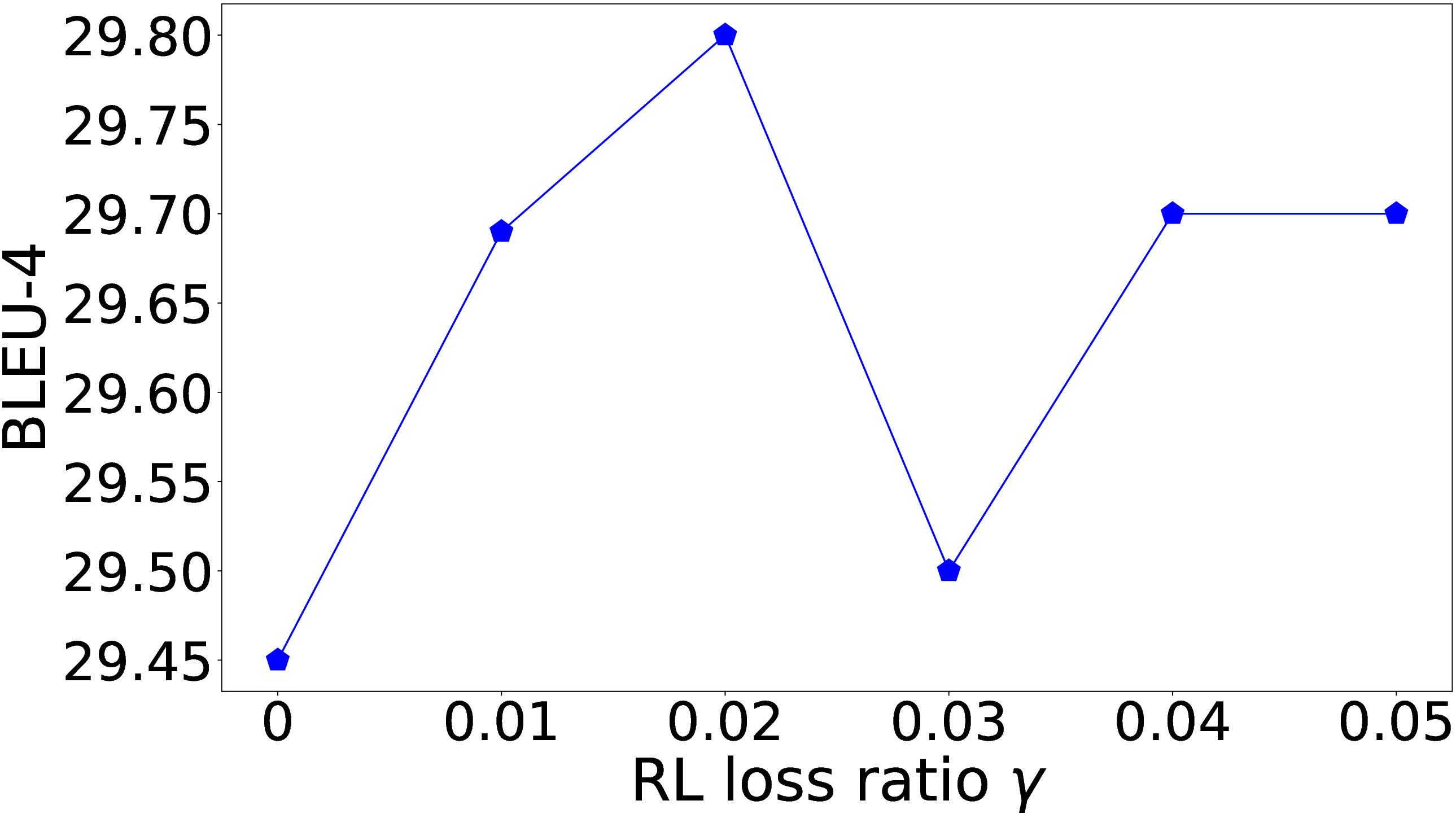}
  \caption{Effect of RL ratio for G2S+AE+RL on WQ.}
  \label{fig:rl_ratio}
  \vspace{-2mm}
\end{figure}

\begin{figure}[!htb]
  \vspace{-2mm}
  \centering
    \includegraphics[keepaspectratio=true,scale=0.15]{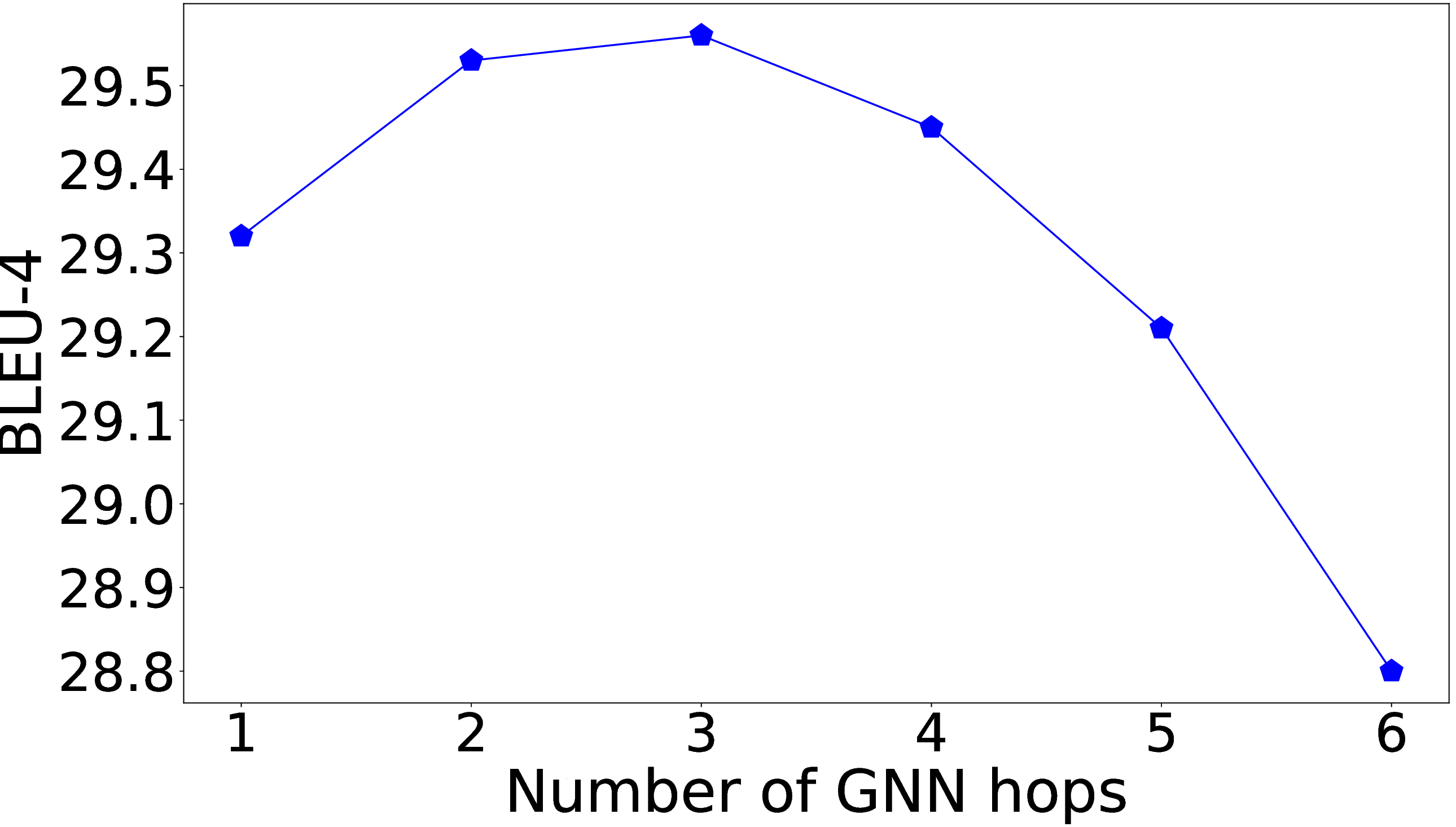}
  \caption{Effect of number of GNN hops for G2S+AE on WQ.}
  \label{fig:gnn_hops}
  \vspace{-1mm}
\end{figure}

\begin{figure}[!htb]
  \vspace{-2mm}
  \centering
   \includegraphics[keepaspectratio=true,scale=0.15]{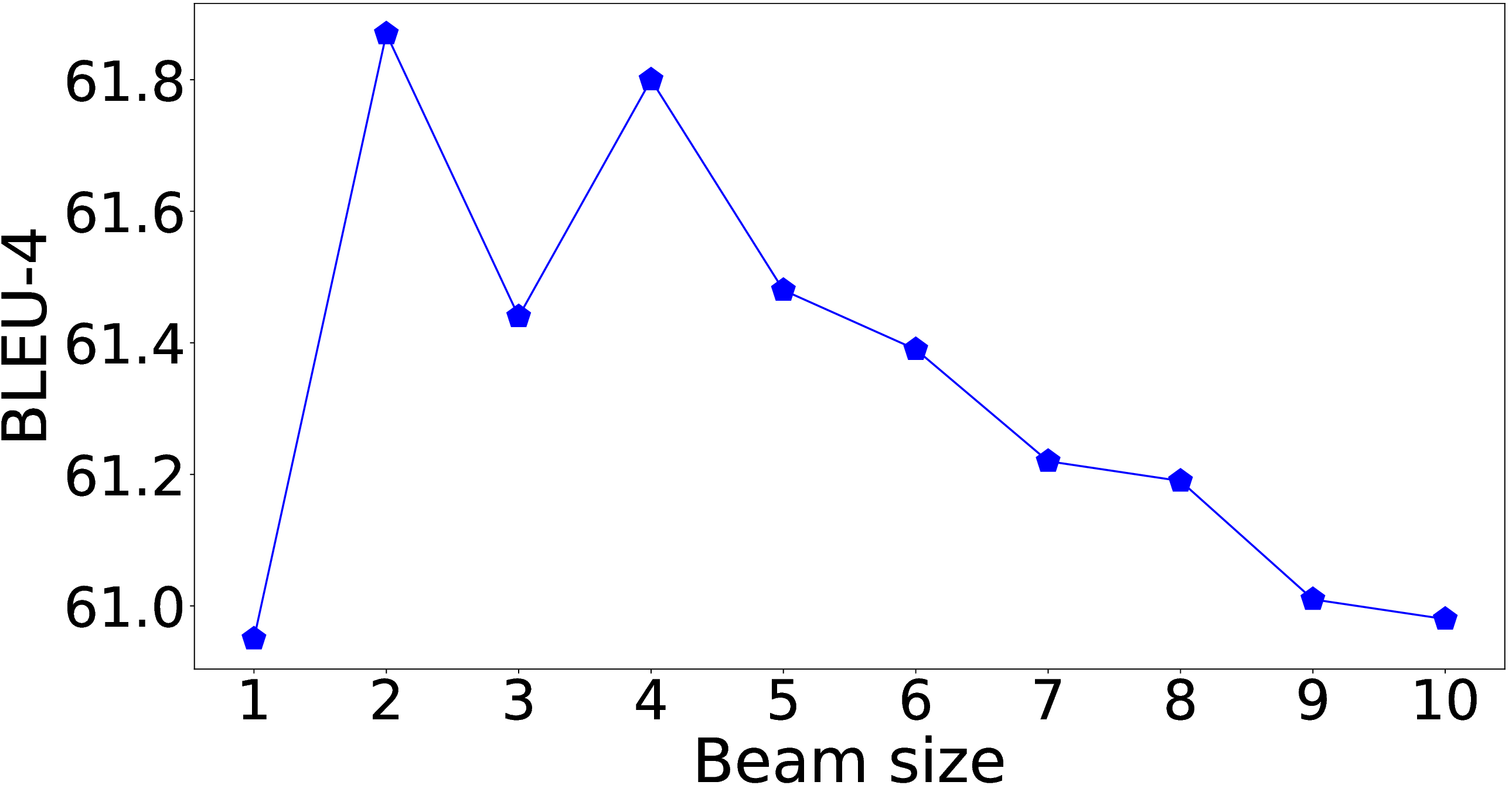}
  \caption{Effect of beam search size for G2S+AE on PQ.}
  \label{fig:beam_size}
  \vspace{-2mm}
\end{figure}

\begin{figure}[!htb]
  \vspace{-1mm}
  \centering
   \includegraphics[keepaspectratio=true,scale=0.1]{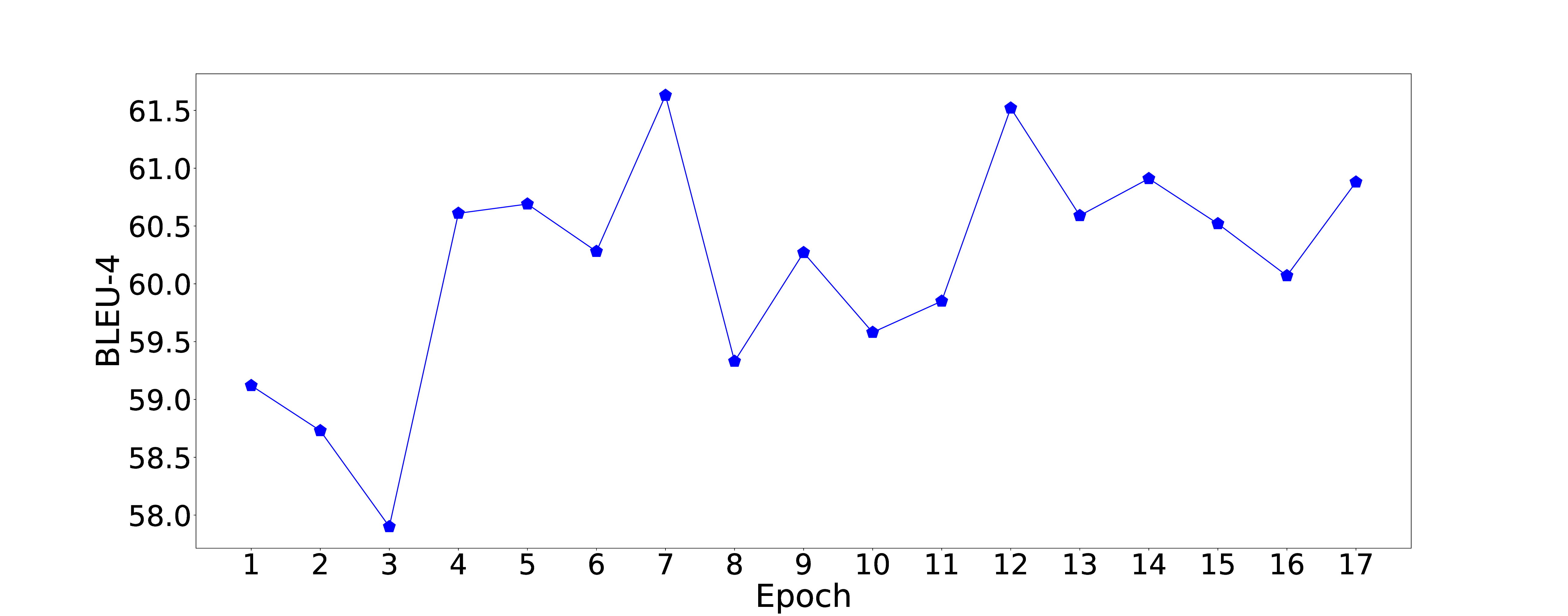}
  \caption{Convergence analysis for G2S+AE on PQ.}
  \label{fig:convergence}
  \vspace{-4mm}
\end{figure}

\subsubsection{Impact of Directionality on GNN Encoder}
As show in~\cref{table:directionality_gnn_results},
we compare the performance of bidirectional Graph2Seq with unidirectional 
(i.e., forward and backward) 
Graph2Seq.
We observe that utilizing the edge direction information in the KG subgraph via bidirectional GNNs can significantly improve the model performance.

\subsubsection{Results on the Two-stage Training Strategy}
\cref{table:wq_rl_results} and \cref{table:pq_rl_results} show the results of training our proposed G2S+AE model with a hybrid objective combining both cross-entropy loss and RL loss following the two-stage training strategy. We denote this variant as G2S+AE+RL.
While the RL-based training strategy boosts the model performance on WQ, it does not help the model training on PQ.
We suspect this is because the PQ dataset is easier compared to the WQ dataset, therefore the benefit of RL-based finetuning on PQ is less significant. 
In order to study how the RL ratio $\gamma$ affects the model performance, we report the test BLEU-4 scores on WQ corresponding to different values of $\gamma$, as shown in~\cref{fig:rl_ratio}. As we can see, compared to $\gamma = 0$ which means no RL-based finetuning is applied, increasing the value of $\gamma$ can help the model performance until certain point.

\subsubsection{Effect of the Number of GNN Hops}
~\cref{fig:gnn_hops} shows the impact of the number of GNN hops when applying a GNN-based encoder to encode the KG subgraph in WQ.
It indicates that increasing the number of GNN hops can boost the model performance until some optimal value.

\subsubsection{Effect of the Beam Search Size}

~\cref{fig:beam_size} shows the impact of the beam size when applying beam search decoding during the testing phase on PQ. It indicates that beam search decoding significantly outperforms greedy search decoding (i.e., beam size = 1) and increasing the beam size can boost the model performance until some optimal value.

\subsubsection{Convergence Analysis}

~\cref{fig:convergence} shows the changes of validation BLEU-4 scores over training epochs on PQ. As we can see, the model was able to converge quickly and achieved the best validation BLEU-4 score after epoch 7.

\subsection{Case Study}

\begin{table}[!htb]
\vspace{-2mm}
\caption{Generated questions on WQ test set. Target answers are underlined. 
For the sake of brevity, we only display the lowest level of the predicate hierarchy.
}
\label{table:case_study}
\centering
\scalebox{1}{
\begin{tabular}{l}
\hline
\textbf{KG subgraph:} 
(\underline{Egypt}, administrative\_divisions, \\Cairo), (Giza Necropolis, contained by, \underline{Egypt})\\

\textbf{Gold:} what country has the city of cairo and \\is home of giza necropolis ?\\

\textbf{G2S w/ KG emb.:} what country that contains \\cairo has cairo as its province      ?\\

\textbf{G2S w/o copy:}  where is the giza giza located\\ in  that has cairo ?\\

\textbf{G2S:} where is the giza necropolis located in\\ that contains cairo ?\\

\textbf{G2S+AE:} what country that contains cairo is \\the location of giza necropolis ?\\
\hline
\end{tabular}
}

\vspace{-3mm}
\end{table}

As shown in~\cref{table:case_study},
we conducted a case study to examine the quality of generated questions using different ablated systems.
First of all, by initializing node/edge embeddings with KG embeddings, the model fails to generate reasonable questions.
As we discussed in~\cref{sec:node_edge_emb}, this might be because of the semantic gap between KG embeddings on the encoder side and word embeddings on the decoder side.
Besides,
with the node-level copying mechanism,
the model was able to directly copy the entity name ``giza necropolis'' from the input KG subgraph into the output question.
Last, incorporating the answer information helps generate more relevant and specific questions. For instance, given the target answer ``Egypt'', the model was able to produce a more specific question which is specifically asking for ``what country'' instead of ``where''.

\subsection{Error Analysis}

\begin{table}[!htb]
\vspace{-2mm}
\caption{Error analysis on generated questions on WQ test set. Target answers are underlined. 
For the sake of brevity, we only display the lowest level of the predicate hierarchy.
}
\label{table:error_analysis}
\centering
\scalebox{1}{
\begin{tabular}{l}
\hline
\textbf{KG subgraph:} (\underline{martin luther king , jr .}, speeches or\\ presentations, /m/05r7ddy), (\underline{martin luther king , jr .}, \\profession, writer), (\underline{martin luther king , jr .},\\ profession, minister of religion), (\underline{martin luther king ,}\\ \underline{jr .}, profession, civil rights activist), (/m/05r7ddy,\\ event, march on washington for jobs and freedom)\\

\textbf{Gold:} who was the speaker at march on washington\\ for jobs and freedom facts ?\\

\textbf{G2S+AE:} who was the speaker in the march on\\ washington for jobs and freedom ?\\

\hline
\hline

\textbf{KG subgraph:} 
(family guy, theme\_song, family guy\\ theme song), (family guy, regular cast, /m/02ntr0s),\\ (/m/02ntr0s, actor, \underline{alex borstein}),
(/m/02ntr0s,\\ character, lois griffin), (/m/02ntr0s, special\\ performance type, voice)\\

\textbf{Gold:} who ' s the voice of stewie griffin from the tv\\ program , with the family guy theme song ?\\

\textbf{G2S+AE:} who is the voice of the voice of the tv\\ program with the family guy family guy theme song \\family guy theme song\\

\hline


\hline
\end{tabular}
}
\vspace{-3mm}
\end{table}

\cref{table:error_analysis} shows some failure cases of our proposed G2S+AE model on the WQ test set. One common syntactic error pattern we observed is repeated words (e.g., repeated ``the voice of'' in the second example) in generated questions. 
Another error pattern is missing important pieces of information. For instance, in the second example\footnote{In this example, the ground-truth question refers to the entity ``stewie griffin'' which is not included in the given input KG subgraph.}, our model failed to utilize the tuple (/m/02ntr0s, character, lois griffin) when generating the question.
The coverage mechanism~\cite{tu2016modeling} is widely used in Seq2Seq models to encourage the full utilization of different tokens in the input text and penalize generating repetitive text. However, in our experiments, we found applying the coverage mechanism did not help improve the overall evaluation scores.
We conjecture this might be because the coverage mechanism can also be too aggressive by encouraging the model to utilize irrelevant tuples in the input KG subgraph.

\subsection{Visualization of the Generated Questions}

\begin{figure}[!htb]
  \vspace{-4mm}
  \centering
    \includegraphics[keepaspectratio=true,scale=0.11]{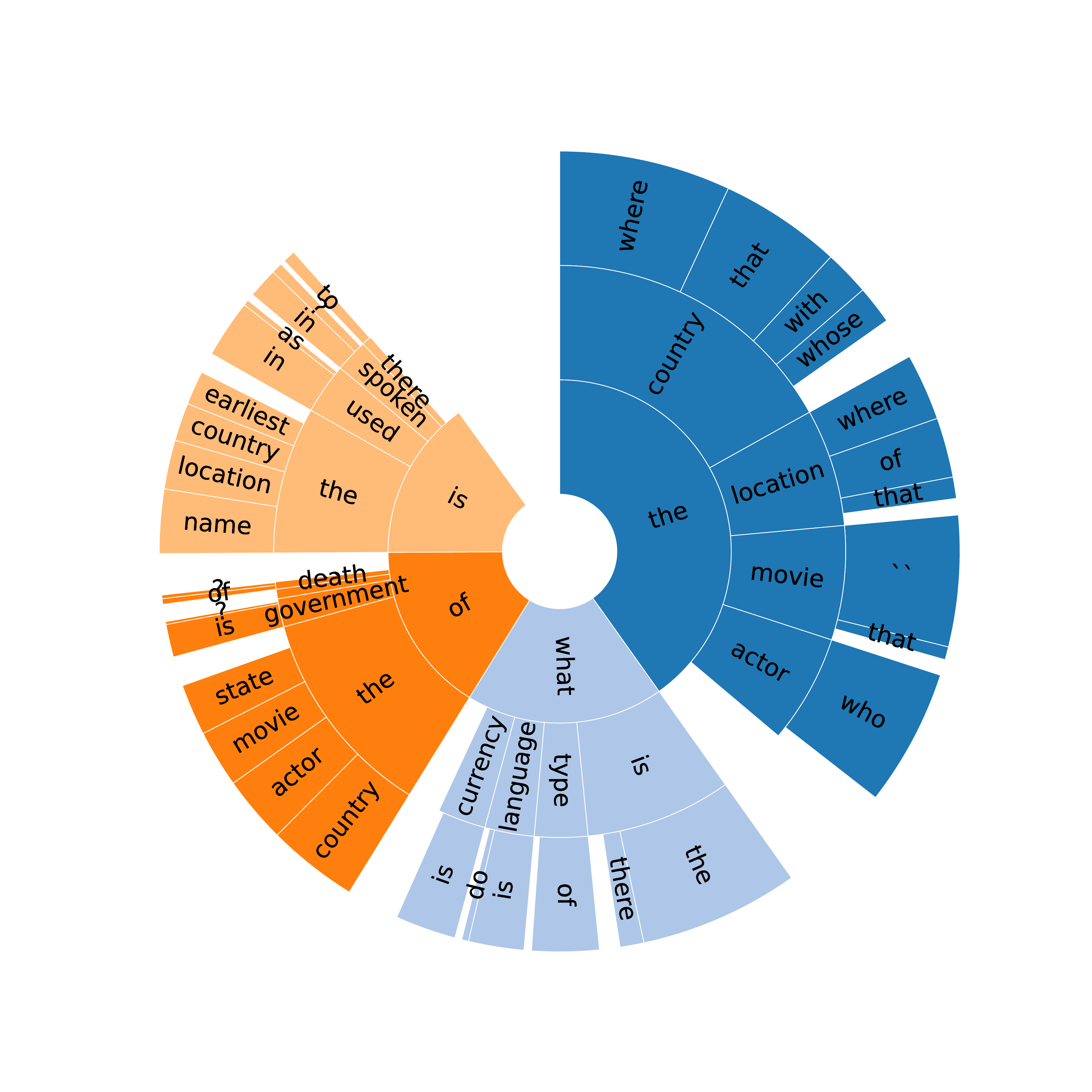}
  \caption{Distribution of trigram prefixes of questions generated by G2S+AE on the WQ test set.}
  \label{fig:pred_vis_q}
  \vspace{-4mm}
\end{figure}

\begin{figure}[!htb]
\vspace{-4mm}
  \centering
    \includegraphics[keepaspectratio=true,scale=0.11]{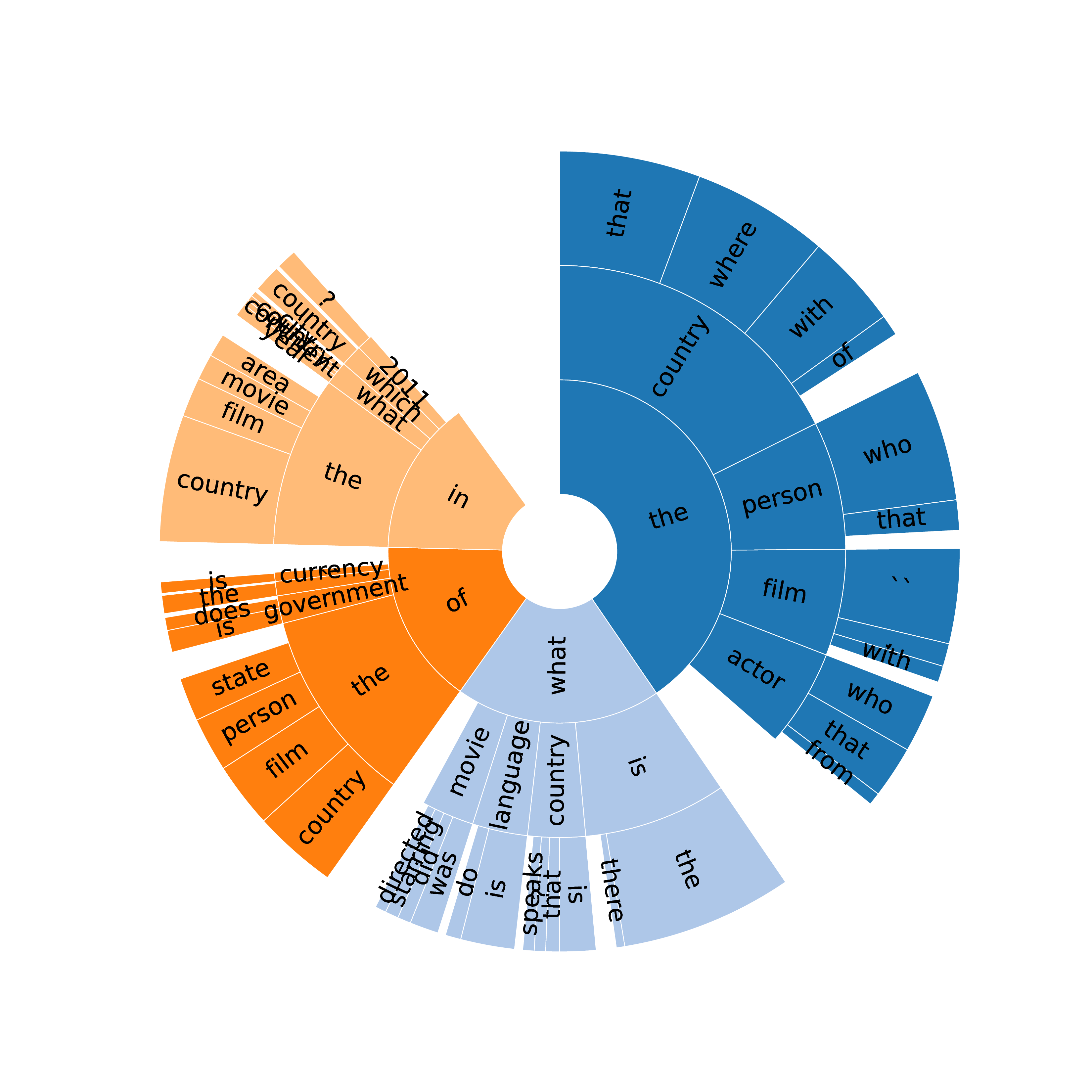}
  \caption{Distribution of trigram prefixes of golden questions in the WQ test set.}
  \label{fig:gold_vis_q}
  \vspace{-5mm}
\end{figure}

\cref{fig:pred_vis_q} and \cref{fig:gold_vis_q} show the distributions of frequent trigram prefixes (i.e., frequency less than 5 not included) of the generated questions and golden questions on the WQ test set. As we can see, our G2S+AE model was able to generate diverse questions which have a similar distribution of trigram prefixes in comparison with the golden questions.

\subsection{QG-Driven Data Augmentation for QA}

\begin{figure}[!htb]
\vspace{-1mm}
  \centering
    \includegraphics[keepaspectratio=true,scale=0.13]{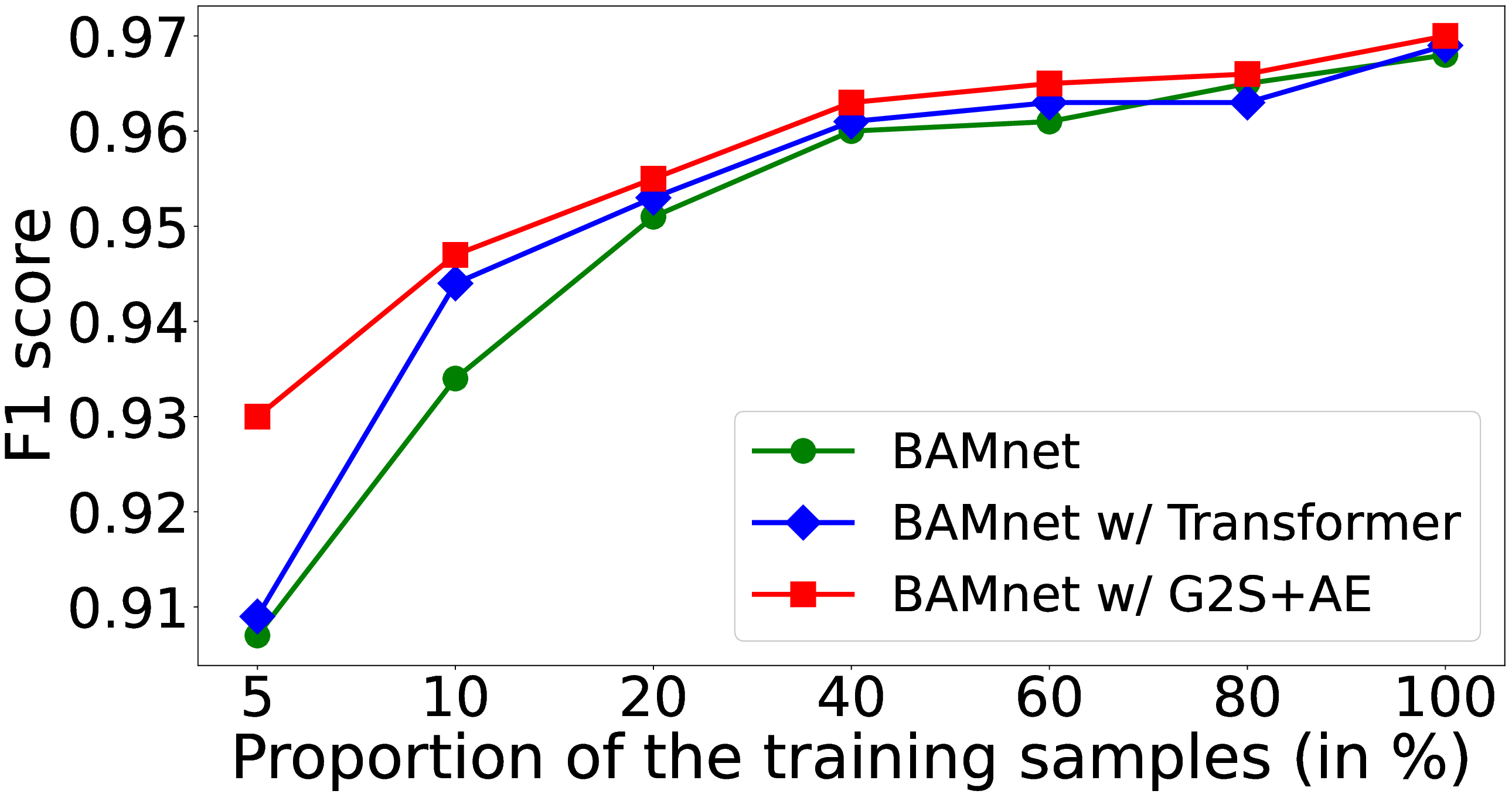}
  \caption{Performance of QG-driven KBQA baseline under different proportions of training data.}
  \label{fig:qa_results}
  \vspace{-3mm}
\end{figure}

One of the most important applications of QG is to generate more training data for QA tasks. 
In this section, we use our proposed QG model to generate more questions for training KBQA methods.
We use WQ as our KBQA benchmark, and randomly split it to 40\%/20\%/40\% (train/dev/test) examples.
As for the KBQA baseline, we use the state-of-the-art KBQA model called BAMnet~\cite{chen2019bidirectional} which directly retrieves answers from a KG by mapping questions and candidate answers into a joint embedding space.
In order to examine the effect of QG-driven data augmentation on the KBQA task, 
we compare the BAMnet baseline with its two data augmentation variants, namely, BAMnet w/ Transformer and BAMnet w/ G2S+AE.
More specifically, the BAMnet baseline is trained only on the part 
(i.e., x\% of the whole training data) 
where gold questions are available, while the other two variants are trained on the combination of the gold questions and the questions automatically generated by two QG models.
Each x\% corresponds to a data point in Fig. 9. We vary the value of x\% from 5\% all the way to 100\% so as to examine the effectiveness of the QG-based data augmentation for KBQA with different training sizes.
Note that given the $x$\% training data, we further randomly split it to 80\%/20\% (train/dev) for training a QG model.

As shown in~\cref{fig:qa_results}, we gradually increase the proportion (i.e., x\%) of the training data, and report the F1 score performance of the above three KBQA model variants.
Here F1 score measures the overlap between the predicted and ground-truth answer set.
The results show that both QG models consistently help improve the KBQA performance when varying x\% training data, and the performance boost is the most significant when training data is scarce (i.e., 5\%, 10\%).
Notably, our G2S+AE model consistently outperforms the Transformer model in improving the KBQA performance.

\section{Conclusion}

In this paper, we introduced a novel bidirectional Graph2Seq model for the KG-QG task.
A novel node-level copying mechanism was proposed to allow directly copying node attributes from the KG subgraph to the output question.
We explored different ways of initializing node/edge embeddings and handling multi-relational graphs.
Our model outperforms existing methods by a significant margin on two benchmarks.

In our experiments, we observed that node/edge embedding initialization has a big impact on the overall model performance. 
We would like to explore more effective ways of initializing node/edge embeddings in the future.
Besides, how to effectively utilize the answer information is critical for generating relevant and meaningful questions. In this work, we introduced simple markup vectors to indicate whether an entity is a target answer or not. We leave more effective ways of answer utilization as future work.
It's also beneficial to design more effective mechanisms to penalize generating repetitive text and encourage fully utilizing important information in the input KG subgraph.
Another interesting direction is to integrate the QG model with KG completion systems. We expect this can be extremely beneficial when the input KG is incomplete, and can potentially lead to generating more interesting and diverse questions.

\section*{Acknowledgments}

The authors thank the editors and reviewers for their constructive feedback.

\bibliographystyle{IEEEtran}
\bibliography{tnnls_2022}

\begin{thebibliography}{10}
\providecommand{\url}[1]{#1}
\csname url@samestyle\endcsname
\providecommand{\newblock}{\relax}
\providecommand{\bibinfo}[2]{#2}
\providecommand{\BIBentrySTDinterwordspacing}{\spaceskip=0pt\relax}
\providecommand{\BIBentryALTinterwordstretchfactor}{4}
\providecommand{\BIBentryALTinterwordspacing}{\spaceskip=\fontdimen2\font plus
\BIBentryALTinterwordstretchfactor\fontdimen3\font minus
  \fontdimen4\font\relax}
\providecommand{\BIBforeignlanguage}[2]{{%
\expandafter\ifx\csname l@#1\endcsname\relax
\typeout{** WARNING: IEEEtran.bst: No hyphenation pattern has been}%
\typeout{** loaded for the language `#1'. Using the pattern for}%
\typeout{** the default language instead.}%
\else
\language=\csname l@#1\endcsname
\fi
#2}}
\providecommand{\BIBdecl}{\relax}
\BIBdecl

\bibitem{du2017learning}
X.~Du, J.~Shao, and C.~Cardie, ``Learning to ask: Neural question generation
  for reading comprehension,'' \emph{arXiv preprint arXiv:1705.00106}, 2017.

\bibitem{song2018leveraging}
L.~Song, Z.~Wang, W.~Hamza, Y.~Zhang, and D.~Gildea, ``Leveraging context
  information for natural question generation,'' in \emph{Proceedings of the
  2018 Conference of the North American Chapter of the Association for
  Computational Linguistics: Human Language Technologies, Volume 2 (Short
  Papers)}, 2018, pp. 569--574.

\bibitem{chen2019reinforcement}
Y.~Chen, L.~Wu, and M.~J. Zaki, ``Reinforcement learning based
  graph-to-sequence model for natural question generation,'' \emph{ICLR}, 2020.

\bibitem{pan2020semantic}
L.~Pan, Y.~Xie, Y.~Feng, T.-S. Chua, and M.-Y. Kan, ``Semantic graphs for
  generating deep questions,'' \emph{arXiv preprint arXiv:2004.12704}, 2020.

\bibitem{li2018visual}
Y.~Li, N.~Duan, B.~Zhou, X.~Chu, W.~Ouyang, X.~Wang, and M.~Zhou, ``Visual
  question generation as dual task of visual question answering,'' in
  \emph{Proceedings of the IEEE Conference on Computer Vision and Pattern
  Recognition}, 2018, pp. 6116--6124.

\bibitem{bao2018table}
J.~Bao, D.~Tang, N.~Duan, Z.~Yan, Y.~Lv, M.~Zhou, and T.~Zhao, ``Table-to-text:
  Describing table region with natural language,'' in \emph{Thirty-Second AAAI
  Conference on Artificial Intelligence}, 2018, pp. 5020--5027.

\bibitem{seyler2017knowledge}
D.~Seyler, M.~Yahya, and K.~Berberich, ``Knowledge questions from knowledge
  graphs,'' in \emph{Proceedings of the ACM SIGIR International Conference on
  Theory of Information Retrieval}.\hskip 1em plus 0.5em minus 0.4em\relax ACM,
  2017, pp. 11--18.

\bibitem{tang2017question}
D.~Tang, N.~Duan, T.~Qin, Z.~Yan, and M.~Zhou, ``Question answering and
  question generation as dual tasks,'' \emph{arXiv preprint arXiv:1706.02027},
  2017.

\bibitem{chen2019bidirectional}
Y.~Chen, L.~Wu, and M.~J. Zaki, ``Bidirectional attentive memory networks for
  question answering over knowledge bases,'' \emph{NAACL}, 2019.

\bibitem{chen2021personalized}
Y.~Chen, A.~Subburathinam, C.~Chen, and M.~J. Zaki, ``Personalized food
  recommendation as constrained question answering over a large-scale food
  knowledge graph,'' in \emph{WSDM 2021}, 2021, pp. 544--552.

\bibitem{heilman2010good}
M.~Heilman and N.~A. Smith, ``Good question! statistical ranking for question
  generation,'' in \emph{Human Language Technologies: The 2010 Annual
  Conference of the North American Chapter of the Association for Computational
  Linguistics}, 2010, pp. 609--617.

\bibitem{mostafazadeh2016generating}
N.~Mostafazadeh, I.~Misra, J.~Devlin, M.~Mitchell, X.~He, and L.~Vanderwende,
  ``Generating natural questions about an image,'' \emph{arXiv preprint
  arXiv:1603.06059}, 2016.

\bibitem{seyler2015generating}
D.~Seyler, M.~Yahya, and K.~Berberich, ``Generating quiz questions from
  knowledge graphs,'' in \emph{Proceedings of the 24th International Conference
  on World Wide Web}.\hskip 1em plus 0.5em minus 0.4em\relax ACM, 2015, pp.
  113--114.

\bibitem{song2016question}
L.~Song and L.~Zhao, ``Question generation from a knowledge base with web
  exploration,'' \emph{arXiv preprint arXiv:1610.03807}, 2016.

\bibitem{vaswani2017attention}
A.~Vaswani, N.~Shazeer, N.~Parmar, J.~Uszkoreit, L.~Jones, A.~N. Gomez,
  {\L}.~Kaiser, and I.~Polosukhin, ``Attention is all you need,'' in
  \emph{Advances in neural information processing systems}, 2017, pp.
  5998--6008.

\bibitem{serban2016generating}
I.~V. Serban, A.~Garc{\'\i}a-Dur{\'a}n, C.~Gulcehre, S.~Ahn, S.~Chandar,
  A.~Courville, and Y.~Bengio, ``Generating factoid questions with recurrent
  neural networks: The 30m factoid question-answer corpus,'' \emph{arXiv
  preprint arXiv:1603.06807}, 2016.

\bibitem{reddy2017generating}
S.~Reddy, D.~Raghu, M.~M. Khapra, and S.~Joshi, ``Generating natural language
  question-answer pairs from a knowledge graph using a rnn based question
  generation model,'' in \emph{Proceedings of the 15th Conference of the
  European Chapter of the Association for Computational Linguistics: Volume 1,
  Long Papers}, 2017, pp. 376--385.

\bibitem{elsahar2018zero}
H.~Elsahar, C.~Gravier, and F.~Laforest, ``Zero-shot question generation from
  knowledge graphs for unseen predicates and entity types,'' \emph{arXiv
  preprint arXiv:1802.06842}, 2018.

\bibitem{liu2019generating}
C.~Liu, K.~Liu, S.~He, Z.~Nie, and J.~Zhao, ``Generating questions for
  knowledge bases via incorporating diversified contexts and answer-aware
  loss,'' in \emph{EMNLP}, 2019, pp. 2431--2441.

\bibitem{hu2022generating}
Y.~Hu, H.~Yang, G.~Zhou, and J.~X. Huang, ``Generating factoid questions with
  question type enhanced representation and attention-based copy mechanism,''
  \emph{Transactions on Asian and Low-Resource Language Information
  Processing}, vol.~21, no.~2, pp. 1--18, 2022.

\bibitem{sutskever2014sequence}
I.~Sutskever, O.~Vinyals, and Q.~Le, ``Sequence to sequence learning with
  neural networks,'' in \emph{Advances in Neural Information Processing
  Systems}, 2014, pp. 3104--3112.

\bibitem{cho2014learning}
K.~Cho, B.~van Merrienboer, C.~Gulcehre, D.~Bahdanau, F.~Bougares, H.~Schwenk,
  and Y.~Bengio, ``Learning phrase representations using rnn encoder--decoder
  for statistical machine translation,'' in \emph{EMNLP}, 2014, pp. 1724--1734.

\bibitem{kumar2019difficulty}
V.~Kumar, Y.~Hua, G.~Ramakrishnan, G.~Qi, L.~Gao, and Y.-F. Li,
  ``Difficulty-controllable multi-hop question generation from knowledge
  graphs,'' in \emph{International Semantic Web Conference}.\hskip 1em plus
  0.5em minus 0.4em\relax Springer, 2019, pp. 382--398.

\bibitem{sheng2020knowledge}
S.~Bi, X.~Cheng, Y.~Li, Y.~Wang, and G.~Qi, ``Knowledge-enriched,
  type-constrained and grammar-guided question generation over knowledge
  bases,'' in \emph{COLING 2020}, 2020, pp. 2776--2786.

\bibitem{hochreiter1997long}
S.~Hochreiter and J.~Schmidhuber, ``Long short-term memory,'' \emph{Neural
  computation}, vol.~9, no.~8, pp. 1735--1780, 1997.

\bibitem{shen2022diversified}
X.~Shen, J.~Chen, J.~Chen, C.~Zeng, and Y.~Xiao, ``Diversified query generation
  guided by knowledge graph,'' in \emph{Proceedings of the Fifteenth ACM
  International Conference on Web Search and Data Mining}, 2022, pp. 897--907.

\bibitem{chen2020kgpt}
W.~Chen, Y.~Su, X.~Yan, and W.~Y. Wang, ``{KGPT:} knowledge-grounded
  pre-training for data-to-text generation,'' in \emph{EMNLP 2020}, 2020, pp.
  8635--8648.

\bibitem{ke2021joint}
P.~Ke, H.~Ji, Y.~Ran, X.~Cui, L.~Wang, L.~Song, X.~Zhu, and M.~Huang,
  ``Jointgt: Graph-text joint representation learning for text generation from
  knowledge graphs,'' in \emph{Findings of the Association for Computational
  Linguistics: {ACL/IJCNLP} 2021}, 2021, pp. 2526--2538.

\bibitem{kipf2016semi}
T.~N. Kipf and M.~Welling, ``Semi-supervised classification with graph
  convolutional networks,'' \emph{arXiv preprint arXiv:1609.02907}, 2016.

\bibitem{gilmer2017neural}
J.~Gilmer, S.~S. Schoenholz, P.~F. Riley, O.~Vinyals, and G.~E. Dahl, ``Neural
  message passing for quantum chemistry,'' in \emph{Proceedings of the 34th
  International Conference on Machine Learning-Volume 70}.\hskip 1em plus 0.5em
  minus 0.4em\relax JMLR. org, 2017, pp. 1263--1272.

\bibitem{hamilton2017inductive}
W.~Hamilton, Z.~Ying, and J.~Leskovec, ``Inductive representation learning on
  large graphs,'' in \emph{Advances in Neural Information Processing Systems},
  2017, pp. 1024--1034.

\bibitem{li2015gated}
Y.~Li, D.~Tarlow, M.~Brockschmidt, and R.~Zemel, ``Gated graph sequence neural
  networks,'' \emph{arXiv preprint arXiv:1511.05493}, 2015.

\bibitem{chen2020iterative}
Y.~Chen, L.~Wu, and M.~J. Zaki, ``Iterative deep graph learning for graph
  neural networks: Better and robust node embeddings,'' in \emph{Advances in
  Neural Information Processing Systems}, 2020, pp. 19\,314--19\,326.

\bibitem{liu2022compact}
N.~Liu, X.~Wang, L.~Wu, Y.~Chen, X.~Guo, and C.~Shi, ``Compact graph structure
  learning via mutual information compression,'' in \emph{TheWebConf 2022},
  2022.

\bibitem{bastings2017graph}
J.~Bastings, I.~Titov, W.~Aziz, D.~Marcheggiani, and K.~Sima'an, ``Graph
  convolutional encoders for syntax-aware neural machine translation,''
  \emph{arXiv preprint arXiv:1704.04675}, 2017.

\bibitem{song2018graph}
L.~Song, Y.~Zhang, Z.~Wang, and D.~Gildea, ``A graph-to-sequence model for
  amr-to-text generation,'' \emph{arXiv preprint arXiv:1805.02473}, 2018.

\bibitem{chen2020graphflow}
Y.~Chen, L.~Wu, and M.~J. Zaki, ``Graphflow: Exploiting conversation flow with
  graph neural networks for conversational machine comprehension,'' in
  \emph{IJCAI 2020}, 2020, pp. 1230--1236.

\bibitem{wu2021graph}
L.~Wu, Y.~Chen, K.~Shen, X.~Guo, H.~Gao, S.~Li, J.~Pei, B.~Long \emph{et~al.},
  ``Graph neural networks for natural language processing: A survey,''
  \emph{Foundations and Trends{\textregistered} in Machine Learning}, vol.~16,
  no.~2, pp. 119--328, 2023.

\bibitem{wu2021deep}
L.~Wu, Y.~Chen, H.~Ji, and B.~Liu, ``Deep learning on graphs for natural
  language processing,'' in \emph{Proceedings of the 44th International ACM
  SIGIR Conference on Research and Development in Information Retrieval}, 2021,
  pp. 2651--2653.

\bibitem{beck2018graph}
D.~Beck, G.~Haffari, and T.~Cohn, ``Graph-to-sequence learning using gated
  graph neural networks,'' \emph{arXiv preprint arXiv:1806.09835}, 2018.

\bibitem{xu2018exploiting}
K.~Xu, L.~Wu, Z.~Wang, M.~Yu, L.~Chen, and V.~Sheinin, ``Exploiting rich
  syntactic information for semantic parsing with graph-to-sequence model,''
  \emph{arXiv preprint arXiv:1808.07624}, 2018.

\bibitem{liu2021retrieval}
S.~Liu, Y.~Chen, X.~Xie, J.~K. Siow, and Y.~Liu, ``Retrieval-augmented
  generation for code summarization via hybrid {GNN},'' in \emph{ICLR 2021},
  2021.

\bibitem{xu2018graph2seq}
K.~Xu, L.~Wu, Z.~Wang, and V.~Sheinin, ``Graph2seq: Graph to sequence learning
  with attention-based neural networks,'' \emph{arXiv preprint
  arXiv:1804.00823}, 2018.

\bibitem{xu2018sql}
K.~Xu, L.~Wu, Z.~Wang, M.~Yu, L.~Chen, and V.~Sheinin, ``Sql-to-text generation
  with graph-to-sequence model,'' \emph{arXiv preprint arXiv:1809.05255}, 2018.

\bibitem{marcheggiani2018deep}
D.~Marcheggiani and L.~Perez-Beltrachini, ``Deep graph convolutional encoders
  for structured data to text generation,'' \emph{arXiv preprint
  arXiv:1810.09995}, 2018.

\bibitem{vougiouklis2018neural}
P.~Vougiouklis, H.~Elsahar, L.-A. Kaffee, C.~Gravier, F.~Laforest, J.~Hare, and
  E.~Simperl, ``Neural wikipedian: Generating textual summaries from knowledge
  base triples,'' \emph{Journal of Web Semantics}, vol.~52, pp. 1--15, 2018.

\bibitem{bordes2013translating}
A.~Bordes, N.~Usunier, A.~Garcia-Duran, J.~Weston, and O.~Yakhnenko,
  ``Translating embeddings for modeling multi-relational data,'' in
  \emph{Advances in neural information processing systems}, 2013, pp.
  2787--2795.

\bibitem{pennington2014glove}
J.~Pennington, R.~Socher, and C.~Manning, ``Glove: Global vectors for word
  representation,'' in \emph{EMNLP}, 2014, pp. 1532--1543.

\bibitem{haussmann2019foodkg}
S.~Haussmann, O.~Seneviratne, Y.~Chen, Y.~Ne’eman, J.~Codella, C.-H. Chen,
  D.~L. McGuinness, and M.~J. Zaki, ``Foodkg: A semantics-driven knowledge
  graph for food recommendation,'' in \emph{International Semantic Web
  Conference}.\hskip 1em plus 0.5em minus 0.4em\relax Springer, 2019, pp.
  146--162.

\bibitem{velivckovic2017graph}
P.~Veli{\v{c}}kovi{\'c}, G.~Cucurull, A.~Casanova, A.~Romero, P.~Lio, and
  Y.~Bengio, ``Graph attention networks,'' \emph{arXiv preprint
  arXiv:1710.10903}, 2017.

\bibitem{ribeiro2019enhancing}
L.~F. Ribeiro, C.~Gardent, and I.~Gurevych, ``Enhancing amr-to-text generation
  with dual graph representations,'' \emph{arXiv preprint arXiv:1909.00352},
  2019.

\bibitem{simonovsky2017dynamic}
M.~Simonovsky and N.~Komodakis, ``Dynamic edge-conditioned filters in
  convolutional neural networks on graphs,'' in \emph{Proceedings of the IEEE
  conference on computer vision and pattern recognition}, 2017, pp. 3693--3702.

\bibitem{levi1942finite}
F.~W. Levi, \emph{Finite geometrical systems: six public lectues delivered in
  February, 1940, at the University of Calcutta}.\hskip 1em plus 0.5em minus
  0.4em\relax The University of Calcutta, 1942.

\bibitem{nair2010rectified}
V.~Nair and G.~E. Hinton, ``Rectified linear units improve restricted boltzmann
  machines,'' in \emph{Proceedings of the 27th international conference on
  machine learning (ICML-10)}, 2010, pp. 807--814.

\bibitem{bahdanau2014neural}
D.~Bahdanau, K.~Cho, and Y.~Bengio, ``Neural machine translation by jointly
  learning to align and translate,'' \emph{arXiv preprint arXiv:1409.0473},
  2014.

\bibitem{luong2015effective}
M.-T. Luong, H.~Pham, and C.~D. Manning, ``Effective approaches to
  attention-based neural machine translation,'' \emph{arXiv preprint
  arXiv:1508.04025}, 2015.

\bibitem{see2017get}
A.~See, P.~J. Liu, and C.~D. Manning, ``Get to the point: Summarization with
  pointer-generator networks,'' \emph{arXiv preprint arXiv:1704.04368}, 2017.

\bibitem{vinyals2015pointer}
O.~Vinyals, M.~Fortunato, and N.~Jaitly, ``Pointer networks,'' in
  \emph{Advances in Neural Information Processing Systems}, 2015, pp.
  2692--2700.

\bibitem{gu2016incorporating}
J.~Gu, Z.~Lu, H.~Li, and V.~O. Li, ``Incorporating copying mechanism in
  sequence-to-sequence learning,'' \emph{arXiv preprint arXiv:1603.06393},
  2016.

\bibitem{koncel2019text}
R.~Koncel-Kedziorski, D.~Bekal, Y.~Luan, M.~Lapata, and H.~Hajishirzi, ``Text
  generation from knowledge graphs with graph transformers,'' \emph{arXiv
  preprint arXiv:1904.02342}, 2019.

\bibitem{bengio2015scheduled}
S.~Bengio, O.~Vinyals, N.~Jaitly, and N.~Shazeer, ``Scheduled sampling for
  sequence prediction with recurrent neural networks,'' in \emph{Advances in
  Neural Information Processing Systems}, 2015, pp. 1171--1179.

\bibitem{ranzato2015sequence}
M.~Ranzato, S.~Chopra, M.~Auli, and W.~Zaremba, ``Sequence level training with
  recurrent neural networks,'' \emph{arXiv preprint arXiv:1511.06732}, 2015.

\bibitem{wu2016google}
Y.~Wu, M.~Schuster, Z.~Chen, Q.~V. Le, M.~Norouzi, W.~Macherey, M.~Krikun,
  Y.~Cao, Q.~Gao, K.~Macherey \emph{et~al.}, ``Google's neural machine
  translation system: Bridging the gap between human and machine translation,''
  \emph{arXiv preprint arXiv:1609.08144}, 2016.

\bibitem{paulus2017deep}
R.~Paulus, C.~Xiong, and R.~Socher, ``A deep reinforced model for abstractive
  summarization,'' \emph{arXiv preprint arXiv:1705.04304}, 2017.

\bibitem{williams1992simple}
R.~J. Williams, ``Simple statistical gradient-following algorithms for
  connectionist reinforcement learning,'' \emph{Machine learning}, vol.~8, no.
  3-4, pp. 229--256, 1992.

\bibitem{rennie2017self}
S.~J. Rennie, E.~Marcheret, Y.~Mroueh, J.~Ross, and V.~Goel, ``Self-critical
  sequence training for image captioning,'' in \emph{Proceedings of the IEEE
  Conference on Computer Vision and Pattern Recognition}, 2017, pp. 7008--7024.

\bibitem{klein2017opennmt}
G.~Klein, Y.~Kim, Y.~Deng, J.~Senellart, and A.~M. Rush, ``Opennmt: Open-source
  toolkit for neural machine translation,'' \emph{arXiv preprint
  arXiv:1701.02810}, 2017.

\bibitem{freebase:datadumps}
Google, ``Freebase data dumps,'' \url{https://developers.google.com/freebase},
  2018.

\bibitem{yih2016value}
W.-t. Yih, M.~Richardson, C.~Meek, M.-W. Chang, and J.~Suh, ``The value of
  semantic parse labeling for knowledge base question answering,'' in
  \emph{Proceedings of the 54th Annual Meeting of the Association for
  Computational Linguistics (Volume 2: Short Papers)}, 2016, pp. 201--206.

\bibitem{talmor2018web}
A.~Talmor and J.~Berant, ``The web as a knowledge-base for answering complex
  questions,'' \emph{arXiv preprint arXiv:1803.06643}, 2018.

\bibitem{zhou2018interpretable}
M.~Zhou, M.~Huang, and X.~Zhu, ``An interpretable reasoning network for
  multi-relation question answering,'' \emph{arXiv preprint arXiv:1801.04726},
  2018.

\bibitem{papineni2002bleu}
K.~Papineni, S.~Roukos, T.~Ward, and W.-J. Zhu, ``Bleu: a method for automatic
  evaluation of machine translation,'' in \emph{Proceedings of the 40th annual
  meeting on association for computational linguistics}.\hskip 1em plus 0.5em
  minus 0.4em\relax Association for Computational Linguistics, 2002, pp.
  311--318.

\bibitem{banerjee2005meteor}
S.~Banerjee and A.~Lavie, ``Meteor: An automatic metric for mt evaluation with
  improved correlation with human judgments,'' in \emph{Proceedings of the acl
  workshop on intrinsic and extrinsic evaluation measures for machine
  translation and/or summarization}, 2005, pp. 65--72.

\bibitem{lin2004rouge}
C.-Y. Lin, ``{ROUGE}: A package for automatic evaluation of summaries,'' in
  \emph{Text Summarization Branches Out}.\hskip 1em plus 0.5em minus
  0.4em\relax Barcelona, Spain: Association for Computational Linguistics,
  2004, pp. 74--81.

\bibitem{kingma2015variational}
D.~P. Kingma, T.~Salimans, and M.~Welling, ``Variational dropout and the local
  reparameterization trick,'' in \emph{Advances in Neural Information
  Processing Systems}, 2015, pp. 2575--2583.

\bibitem{kingma2014adam}
D.~P. Kingma and J.~Ba, ``Adam: A method for stochastic optimization,''
  \emph{arXiv preprint arXiv:1412.6980}, 2014.

\bibitem{tu2016modeling}
Z.~Tu, Z.~Lu, Y.~Liu, X.~Liu, and H.~Li, ``Modeling coverage for neural machine
  translation,'' \emph{arXiv preprint arXiv:1601.04811}, 2016.

\end{thebibliography}


\end{document}